\documentclass[sigconf]{acmart}
\usepackage{multirow}
\usepackage[normalem]{ulem}
\useunder{\uline}{\ul}{}
\usepackage{subfig}
\AtBeginDocument{%
  }

\setcopyright{acmlicensed}
\copyrightyear{2025}
\acmYear{2025}
\acmDOI{XXXXXXX.XXXXXXX}

\renewcommand\footnotetextcopyrightpermission[1]{}

\acmConference[Conference acronym 'XX]{Make sure to enter the correct
  conference title from your rights confirmation emai}{XX XX--XX,
  20XX}{XX, XX}
\acmISBN{978-1-4503-XXXX-X/18/06}

\usepackage{array}
\usepackage{bbding}
\usepackage{enumitem}
\begin{document}

\title{GraphTool-Instruction: Revolutionizing Graph Reasoning in LLMs through Decomposed Subtask Instruction
}

\author{Rongzheng Wang}
\email{wangrongzheng@std.uestc.edu.cn}
\affiliation{%
  \institution{University of Electronic Science and Technology of China}
  \city{Chengdu}
  \country{China}
}

\author{Shuang Liang*}
\email{shuangliang@uestc.edu.cn}
\affiliation{%
  \institution{University of Electronic Science and Technology of China}
  \city{Chengdu}
  \country{China}
}

\author{Qizhi Chen}
\email{202311081625@std.uestc.edu.cn}
\affiliation{%
  \institution{University of Electronic Science and Technology of China}
  \city{Chengdu}
  \country{China}
}

\author{Jiasheng Zhang}
\email{zjss12358@std.uestc.edu.cn}
\affiliation{%
  \institution{University of Electronic Science and Technology of China}
  \city{Chengdu}
  \country{China}
}

\author{Ke Qin}
\email{qinke@uestc.edu.cn}
\affiliation{%
  \institution{University of Electronic Science and Technology of China}
  \city{Chengdu}
  \country{China}
}

% \affiliation{%

% \affiliation{%

% \renewcommand{\shortauthors}{anonymous}

\begin{abstract}

Large language models (LLMs) have been demonstrated to possess the capabilities to understand fundamental graph properties and address various graph reasoning tasks. Existing methods fine-tune LLMs to understand and execute graph reasoning tasks by specially designed task instructions. However, these Text-Instruction methods generally exhibit poor performance. Inspired by tool learning, researchers propose Tool-Instruction methods to solve various graph problems by special tool calling (\textit{e.g.}, function, API and model), achieving significant improvements in graph reasoning tasks. Nevertheless, current Tool-Instruction approaches focus on the tool information and ignore the graph structure information, which leads to significantly inferior performance on small-scale LLMs (less than 13B). To tackle this issue, we propose GraphTool-Instruction, an innovative Instruction-tuning approach that decomposes the graph reasoning task into three distinct subtasks (\textit{i.e.}, \textit{graph extraction}, \textit{tool name identification} and \textit{tool parameter extraction}), and design specialized instructions for each subtask. 
Our GraphTool-Instruction can be used as a plug-and-play prompt for different LLMs without fine-tuning.
Moreover, building on GraphTool-Instruction, we develop GTools, a dataset that includes twenty graph reasoning tasks, and create a graph reasoning LLM called GraphForge based on Llama3-8B. 
We conduct extensive experiments on twenty graph reasoning tasks with different graph types (\textit{e.g.}, graph size or graph direction), and we find that GraphTool-Instruction achieves SOTA compared to Text-Instruction and Tool-Instruction methods. Fine-tuned on GTools, GraphForge gets further improvement of over 30\% compared to the Tool-Instruction enhanced GPT-3.5-turbo, and it performs comparably to the high-cost GPT-4o. Our codes and data are available at \url{https://anonymous.4open.science/r/GraphTool-Instruction/}.
\end{abstract}





\maketitle

\section{Introduction}
\label{Introduction}
Although Large Language Models (LLMs) excel in fields such as natural language processing, they encounter significant challenges when dealing with graph data. Graph structures exhibit high connectivity, rich combinatorial properties and Non-Euclidean characteristics, making their processing fundamentally different from traditional text or image data~\cite{DBLP:conf/acl/WangKMLSKH23, DBLP:conf/iccv/KirillovMRMRGXW23}. Research~\cite{DBLP:conf/nips/WangFHTHT23} shows that LLMs possess a basic understanding of graph properties and the capabilities to solve graph reasoning tasks, but their accuracy remains considerably deficient because of two challenges.
\begin{table}[t]
\small
\caption{A summary of LLMs-based graph reasoning methods. For conciseness, Function Calling is abbreviated as FC.}
\renewcommand{\arraystretch}{1.3}
\begin{tabular}{cccc}
\hline
\multirow{2}{*}{Method} & \multicolumn{3}{c}{Instruction Level} \\ \cline{2-4} 
 & Text & Tool & GraphTool \\ \hline
Talk like graph~\cite{DBLP:journals/corr/abs-2310-04560} & \Checkmark & \XSolidBrush & \XSolidBrush \\
GraphInstruct~\cite{DBLP:journals/corr/abs-2403-04483} & \Checkmark & \XSolidBrush & \XSolidBrush \\
GraphWiz~\cite{DBLP:journals/corr/abs-2402-16029} & \Checkmark & \XSolidBrush & \XSolidBrush \\
Graph-ToolFormer~\cite{DBLP:journals/corr/abs-2304-11116} & \Checkmark & \Checkmark & \XSolidBrush \\
GPT-4o-FC~\cite{DBLP:journals/corr/abs-2303-08774} & \Checkmark & \Checkmark & \XSolidBrush \\
GLM4-0520-FC~\cite{glm2024chatglm} & \Checkmark & \Checkmark & \XSolidBrush \\
GraphTool-Instruction & \Checkmark & \Checkmark & \Checkmark \\ \hline
\end{tabular}
\label{Table1}
\vspace{-1.5em}
\end{table}

\begin{itemize}[leftmargin=*]
\item \textbf{Challenge 1 Graph Understanding (GU)}: Can LLMs really understand the correct topology information of the graph with natural language?
\item \textbf{Challenge 2 Graph Processing (GP)}: Can LLMs really have the ability to solve the graph algorithm (\textit{e.g.}, Shortest Path Problem) with generative model inference? 
\end{itemize}

\begin{figure}[h]
\centering 
\includegraphics[width=3.3in]
{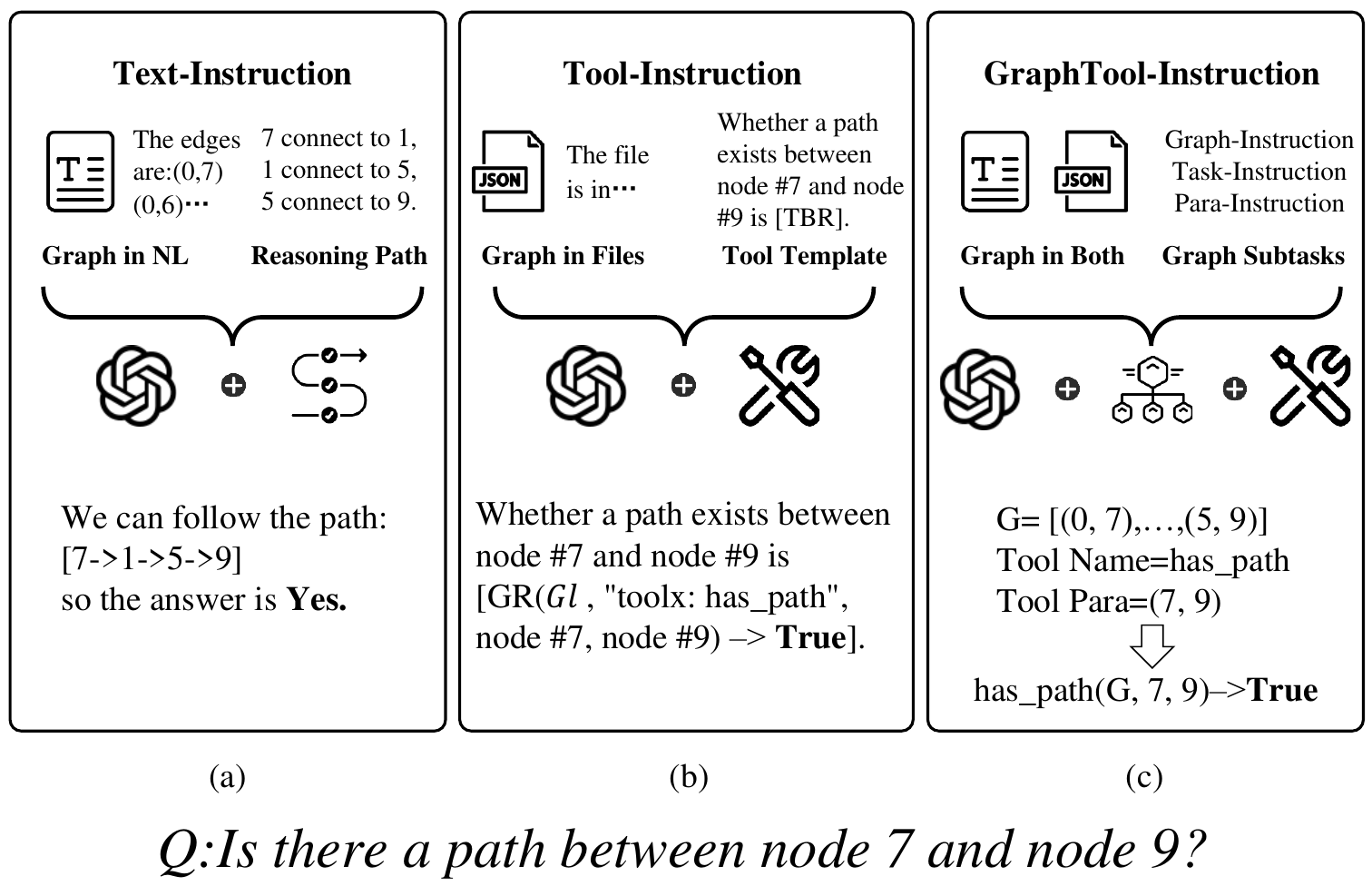}%%%%%%%%%%%%%%%%scale=缩小比例，或者用width=2in
\vspace{-0.5em}
\caption{(a) Text-Instruction method represented by GraphWiz; (b) Tool-Instruction method represented by Graph-ToolFormer; (c) GraphTool Instruction-tuning method.}  
\vspace{-1.0em}
\Description{(a) Text-Instruction method represented by GraphWiz; (b) Tool-Instruction method represented by Graph-ToolFormer; (c) GraphTool Instruction-tuning method.}  
\label{Fig: introduction}
\end{figure}

To enhance the \textbf{GP} ability of LLMs, researchers explore the \textbf{Text-Instruction} approaches which initially begin with the Chain of Thought (CoT)~\cite{DBLP:conf/nips/Wei0SBIXCLZ22}. This method emphasizes that LLMs can solve graph problems through step-by-step reasoning. However, the performance improvement by CoT is limited in \textbf{GU} challenge when dealing with complex graph reasoning tasks. Recent studies have proposed methods that rely on providing additional graph information prompts~\cite{DBLP:journals/corr/abs-2310-04560} or explicit reasoning steps~\cite{DBLP:journals/corr/abs-2403-04483, DBLP:journals/corr/abs-2402-16029} to enhance the LLMs' \textbf{GU}  ability (see Figure~\ref{Fig: introduction} a). While these methods show improvements within their specific prompted tasks, they may compromise their effectiveness in out-of-domain tasks.

Some researchers realize that LLMs cannot deal with all professional problems with only generative model inference.
Toolformer~\cite{DBLP:conf/nips/SchickDDRLHZCS23} first introduces the concept of tool learning into LLMs, which not only generates text but also can call on and utilize specific tools to solve more complex and professional problems.
Inspired by Toolformer, Graph-ToolFormer~\cite{DBLP:journals/corr/abs-2304-11116} is the first work that introduces the \textbf{Tool-Instruction} method to solve \textbf{GP} challenge in graph reasoning tasks (see Figure~\ref{Fig: introduction} b). However, Graph-ToolFormer suffers from \textbf{GU} problem, which shows limited performance on queries containing both the graph information and task description, such as
NLGraph~\cite{DBLP:conf/nips/WangFHTHT23} and GraphInstruct~\cite{DBLP:journals/corr/abs-2403-04483}.
Meanwhile, constrained by issues such as the diversity of tool names and tool calling formats, Graph-ToolFormer exhibits low performance on the \textbf{GP} challenge. 

Current most powerful closed-source LLMs like GPT-4o~\cite{DBLP:journals/corr/abs-2303-08774}, Claude-3-opus~\cite{TheC3} and GLM4-0520~\cite{glm2024chatglm} have introduced a new Tool-Instruction method (\textit{i.e.}, Function Calling). This method allows LLMs to act as controllers that integrate tool descriptions into the input prompts. Function Calling improves LLMs' ability to understand and use external tools, making them more flexible without additional training~\cite{DBLP:journals/corr/abs-2402-10466, DBLP:journals/corr/abs-2405-17438, DBLP:journals/corr/abs-2305-16504, DBLP:conf/nips/LuPCGCWZG23}. However, this method depends on the quality of the tool documentation and output text by LLMs which may result in instances of failure or incorrect tool invocation ~\cite{DBLP:journals/corr/abs-2310-07075}. Especially when dealing with graph reasoning tasks, the internal Function Calling mechanism of LLMs struggles to handle the \textbf{GU} challenge.

In this study, we propose \textbf{GraphTool-Instruction}, a novel method designed to enhance the capabilities of LLMs in addressing graph reasoning tasks. Compared with methods listed in Table~\ref{Table1}, our method first decomposes the graph reasoning task into three subtasks \textit{graph extraction}, \textit{tool name identification} and \textit{parameter extraction}.
For the \textbf{GU} challenge, we propose Graph-Instruction to solve \textit{graph extraction} task, which enhances the LLMs to identify and extract graph structure information from natural language or file paths. For the \textbf{GP} challenge, we decompose traditional Tool-Instruction into Task-Instruction and Parameter-Instruction. At first, Task-Instruction guides LLMs to choose the right graph tools for solving graph reasoning tasks, along with constraints on the output formats of tools.
Then the Parameter-Instruction retrieves the graph tool parameter for tasks that require specific inputs, such as the starting and ending nodes in the Shortest Path task. 
Our experiment results demonstrate that, even without fine-tuning, our approach achieves an average accuracy of 94\% on Llama3-8B~\cite{DBLP:journals/corr/abs-2302-13971}, markedly outperforming Text-Instruction methods over 40\% and GPT-3.5-turbo-FC over 30\%. 
To further enhance the reasoning capabilities of all LLMs, we have developed \textbf{GTools}, an Instruction-tuning dataset that includes twenty different graph reasoning tasks with 40,000 instances. 
Moreover, we develop an open-source LLM for graph reasoning tasks called \textbf{GraphForge} based on Llama3-8B fine-tuned with our proposed GTools.
Our GraphForge achieves an average accuracy of over 98\% on all graph reasoning tasks which performs comparably to the high-cost GPT-4o-FC. The contributions of this paper are as follows:

% \vspace{-1.0em}
\begin{itemize}[leftmargin=*]
\item We summarize LLMs-based graph reasoning methods and propose \textbf{GraphTool-Instruction}, a novel GraphTool level Instruction-tuning method which first decomposes the graph reasoning task into three subtasks \textit{graph extraction}, \textit{tool name identification} and \textit{parameter extraction} with corresponding Graph-Instruction, Task-Instruction and Parameter-Instruction. 
\item We develop \textbf{GTools}, the first GraphTool-Instruction dataset comprising twenty types of graph reasoning tasks. GTools excels in both the variety of tasks and the scale of graphs, thus posing a greater challenge to LLMs in capturing graph structure information. Furthermore, we develop \textbf{GraphForge} based on Llama3-8B fine-tuned with GTools.
\item By incorporating Graph-Instruction and Parameter-Instruction, GraphTool-Instruction significantly improves the accuracy of Tool-Instruction, and achieves state-of-the-art results among all Tool-Instruction methods, except being 1\% behind GPT-4o-FC.
\item We have introduced three new evaluation metrics: \textit{Graph}, \textit{Tool Name} and \textit{Tool Parameter} to enhance the reliability of our dataset. Furthermore, we utilize the accuracy rates of these three metrics to deeply analyze the factors that affect the tool execution results.
\end{itemize}

\begin{figure*}[t]
\centering
\includegraphics[width=1.0\textwidth]{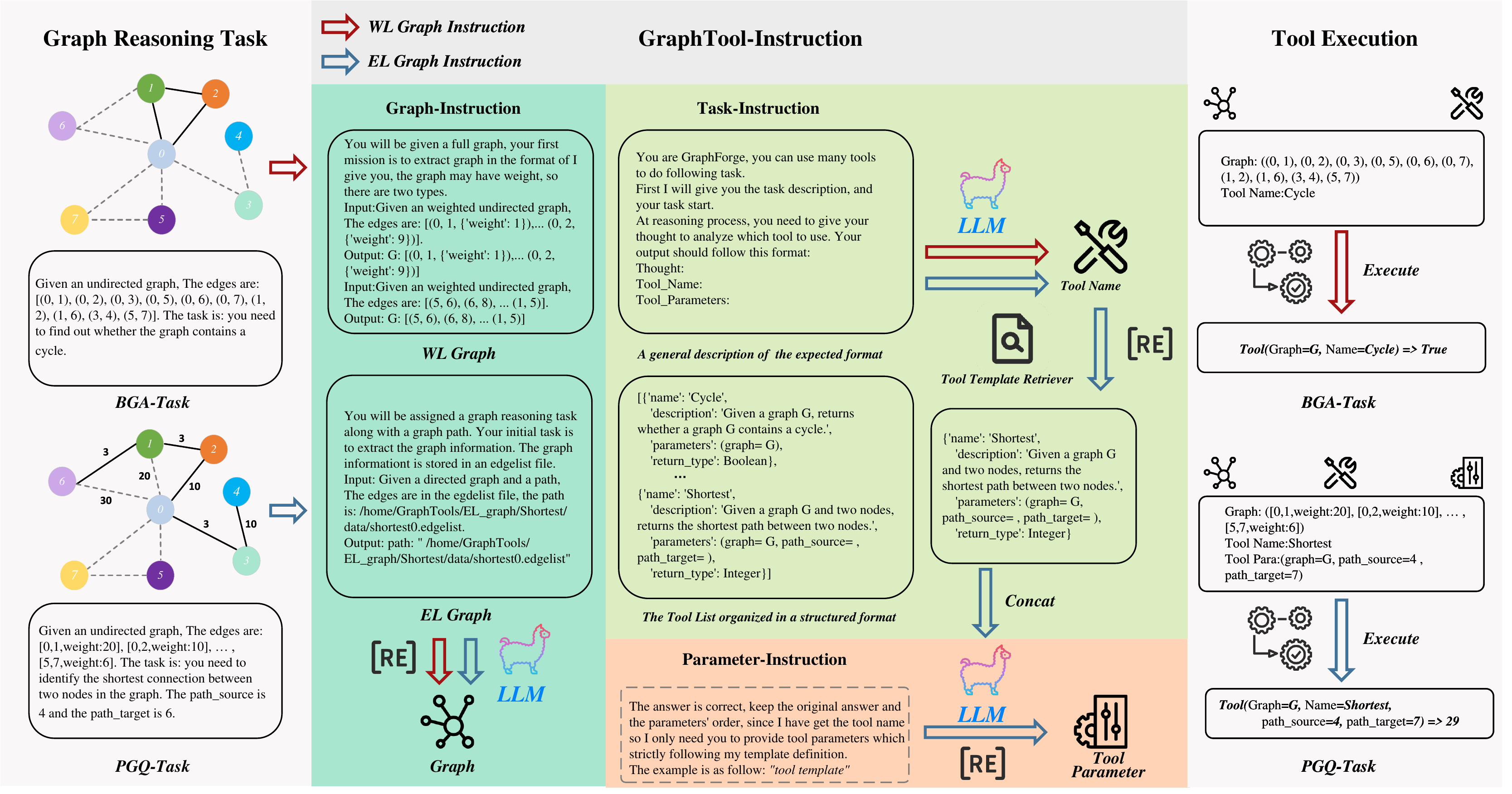}
\caption{The overview of LLM solves graph reasoning tasks based on GraphTool-Instruction. Basic Graph Analysis Task (BGA-Task) does not require additional tool parameters, whereas Parametric Graph Query Task (PGQ-Task) requires specific input tool parameters for reasoning. WL-Graph denotes a task length within 4096 tokens, while EL-Graph is the opposite. The red arrow shows the BGA-Task reasoning process, and the blue arrow shows the PGQ-Task process, which additionally introduces Parameter-Instruction to enhance the accuracy of parameter extraction.
}  
\vspace{-1.0em}
\Description{The overview of LLM solves graph reasoning tasks based on GraphTool-Instruction. Basic Graph Analysis Task (BGA-Task) does not require additional tool parameters, whereas Parametric Graph Query Task (PGQ-Task) requires specific input tool parameters for reasoning. WL-Graph denotes a task length within 4096 tokens, while EL-Graph is the opposite. The red arrow shows the BGA-Task reasoning process, and the blue arrow shows the PGQ-Task process, which additionally introduces Parameter-Instruction to enhance the accuracy of parameter extraction.
}  
\label{Fig: method}
\end{figure*}

\section{Related Works}
Amidst the proliferation of LLMs, there is growing scholarly interest in integrating these computational frameworks with graph data~\cite{DBLP:journals/sigkdd/ChenMLJWWWYFLT23, DBLP:journals/corr/abs-2312-02783, DBLP:journals/corr/abs-2305-13168, DBLP:journals/corr/abs-2312-06185, DBLP:conf/emnlp/JiangZDYZW23}. On the one hand, practical performance assessments are conducted by some researchers, investigating whether LLMs have the capabilities of reasoning on graphs such as NLGraph~\cite{DBLP:conf/nips/WangFHTHT23}, GPT4Graph~\cite{DBLP:journals/corr/abs-2305-15066} and GraphInstruct~\cite{DBLP:journals/corr/abs-2403-04483}. On the other hand, several studies, exemplified by Talk like a Graph~\cite{DBLP:journals/corr/abs-2310-04560}, explore the impact of different graph description languages on LLMs' understanding of graph data. These studies inspire researchers to develop a comprehensive method that enables LLMs to solve various types of graph reasoning tasks.

\subsection{Text-Instruction} 
CoT~\cite{DBLP:conf/nips/Wei0SBIXCLZ22} is always the primary approach for solving graph reasoning tasks. Many researchers have conducted explorations based on this method and its variants~\cite{DBLP:journals/corr/abs-2307-07697, DBLP:journals/corr/abs-2310-17110, DBLP:journals/corr/abs-2403-04483}. Results indicate that CoT-style reasoning can improve performance on simple graph reasoning tasks, such as Cycle Detection and Shortest Path. However, this improvement is inconsistent on more complex tasks, such as Maximum Flow and Topological Sorting~\cite{DBLP:conf/nips/WangFHTHT23, DBLP:journals/corr/abs-2308-11224, DBLP:journals/corr/abs-2305-15066}. To enhance the performance on complex graph reasoning tasks, GraphWiz~\cite{DBLP:journals/corr/abs-2402-16029} employs GPT-4 to generate initial reasoning paths and utilizes a multi-sampling approach to enhance the accuracy of model outputs, demonstrating high accuracy and generalization capabilities across various graph problems. GraphInstruct~\cite{DBLP:journals/corr/abs-2403-04483} enhances reasoning performance by providing a diverse array of graph generation processes and detailed reasoning steps, as well as a step mask training strategy. Although results have demonstrated their effectiveness, these methods exhibit poor performance in out-of-domain tasks.

\subsection{Tool-Instruction} 
The integration of external tools ~\cite{DBLP:conf/nips/SchickDDRLHZCS23, DBLP:conf/iclr/GouSGSYHDC24, DBLP:journals/corr/abs-2401-06201, DBLP:conf/aaai/BestaBKGPGGLNNH24} is a novel way to boost LLMs' functional capabilities. A common approach is to create Instruction-tuning datasets ~\cite{DBLP:journals/corr/abs-2307-16789, DBLP:conf/emnlp/LiZ000YLHL23, DBLP:journals/corr/abs-2305-15334, DBLP:journals/corr/abs-2306-05301, DBLP:journals/corr/abs-2305-16504} which include scenarios requiring tools. The LLMs are then fine-tuned to output text that effectively uses these tools. However, these efforts mainly focus on generating tool calls within real-world scenarios. In fact, graph reasoning tasks are highly suitable for reasoning by tools. Graph-ToolFormer~\cite{DBLP:journals/corr/abs-2304-11116} represents the pioneering research that enables LLMs to generate tool calls for reasoning. This approach is suitable for graph reasoning tasks from knowledge graphs, social networks and recommendation systems. However, the tool calls are executed to acquire answers on an external graph which limits its ability to generalize across graph reasoning tasks organized in natural language.

Despite their potential, the use of Tool-Instruction with the current most powerful closed-source LLMs like GPT-4o~\cite{DBLP:journals/corr/abs-2303-08774}, Claude-3-opus~\cite{TheC3} and GLM4-0520~\cite{glm2024chatglm} is limited because they can't easily add new tools. To address these issues, these LLMs have introduced a new approach called Function Calling. This approach allows LLMs to act as controllers that integrate tool descriptions right into the input prompts, significantly enhancing LLMs' ability to understand and utilize external tools, making them more functional and flexible without additional training. However, this approach depends on the quality of the tool documentation and output text by LLMs. Consequently, output text that is difficult to parse may result in instances of failure or incorrect tool calls ~\cite{DBLP:journals/corr/abs-2310-07075}.

\section{Method}
% In this section, we formally define the GraphTool-Instruction with three components oriented towards three subtasks: \textit{graph extraction}, \textit{tool name identification} and \textit{tool parameter extraction}. We then generate various types of graph reasoning tasks along with their corresponding answer labels. Based on GraphTool-Instruction, we employ Llama3-8B to create candidate answer texts. We extract the necessary information for tool execution and compare the results with the answer labels. The correctly executed answer texts will be retained and paired with GraphTool-Instruction and the corresponding tasks to form our Instruction-tuning dataset, \textit{i.e.}, GTools. A summary of our method is provided in Figure \ref{Fig: method}.

In this section, we formally illustrate the GraphTool-Instruction with three components Graph-Instruction, Task-Instruction and Parameter-Instruction oriented towards three subtasks: \textit{graph extraction}, \textit{tool name identification} and \textit{tool parameter extraction}. Based on GraphTool-Instruction, we employ Llama3-8B to solve graph reasoning tasks and generate candidate answer texts. After extracting the necessary information for tool execution and comparing the results with the answer labels, the correctly executed answer texts will be retained and paired with GraphTool-Instruction and graph tasks to form our Instruction-tuning dataset (\textit{i.e.}, GTools). A brief illustration of our method is provided in Figure \ref{Fig: method}.

\subsection{GraphTool-Instruction Construction}
We select eleven classic graph reasoning tasks and generate examples for both directed and undirected graphs for each task. The list of tasks is presented in Table \ref{Table2} and the detailed descriptions are in \textbf{Appendix~\ref{appendixd Details of the graph reasoning task definition}}. Notably, we define the Maximum Triangle Sum as finding the triangle with the largest sum of edge weights, therefore this task is exclusive to undirected graphs. While Topological Sorting is specific to directed graphs. In total, we identify twenty distinct graph reasoning tasks. For these graph reasoning tasks, we classify them into two primary categories as shown in Figure~\ref{Fig: method}. The first category is \textbf{Basic Graph Analysis Task} (\textbf{BGA-Task}), which generally requires information on the graph structure and the tool name such as Cycle Detection. The second category is \textbf{Parametric Graph Query Task} (\textbf{PGQ-Task}) with the necessity for additional parameter inputs, \textit{e.g.}, the Shortest Path necessitates the starting and ending nodes. 

We propose GraphTool-Instruction in addressing graph reasoning tasks with three components: Graph-Instruction, Task-Instruction and Parameter-Instruction. We employ Graph-Instruction and Task-Instruction as shared components for both BGA-Task and PGQ-Task. Graph-Instruction is designed to extract the graph structure information, whereas Task-Instruction is designed to enable LLMs to identify the tool names and extract parameters associated with these tools. It should be noted that, while the tool names identified from Task-Instruction are highly accurate, the associated tool parameters frequently suffer from issues such as omissions, incorrect order and non-compliance with the required execution formats. Consequently, for PGQ-Tasks, we propose Parameter-Instruction to enable LLMs to output tool parameters in a more accurate and standardized format based on the tool name provided by Task-Instruction. Therefore, these three instructions can be summarized serving three subtasks: \textit{graph extraction}, \textit{tool name identification} and \textit{tool parameter extraction}.

\begin{table}[t]
\small
\caption{Graph reasoning task summary. For conciseness, Topological Sorting and Maximum Triangle Sum are abbreviated as Topo and Triangle, respectively.}
\renewcommand{\arraystretch}{1.2}
\begin{tabular}{ccccc}
\hline
\multicolumn{2}{c}{\multirow{2}{*}{Task}} & \multirow{2}{*}{Weighted} & \multicolumn{2}{c}{Graph Type} \\
\multicolumn{2}{c}{} &  & Directed & Undirected \\ \hline
\multirow{5}{*}{BGQ} & Cycle Detection & \XSolidBrush & \Checkmark & \Checkmark \\
 & Triangle & \Checkmark & \XSolidBrush & \Checkmark \\
 & Edge Count & \XSolidBrush & \Checkmark & \Checkmark \\
 & Node Count & \XSolidBrush & \Checkmark & \Checkmark \\
 & Topo & \XSolidBrush & \Checkmark & \XSolidBrush \\ \cline{2-5} 
\multirow{6}{*}{PGQ} & Degree Count & \XSolidBrush & \Checkmark & \Checkmark \\
 & Edge Existence & \XSolidBrush & \Checkmark & \Checkmark \\
 & Node Existence & \XSolidBrush & \Checkmark & \Checkmark \\
 & Maximum Flow & \Checkmark & \Checkmark & \Checkmark \\
 & Path Existence & \XSolidBrush & \Checkmark & \Checkmark \\
 & Shortest Path & \Checkmark & \Checkmark & \Checkmark \\ \hline
\end{tabular}
\label{Table2}
\vspace{-2.0em}
\end{table}

\textbf{Graph-Instruction (for GU)}: This instruction is designed to enable LLMs to extract graph structure information from the given tasks. To assess and enhance the capabilities of LLMs in processing graphs of various sizes, we establish a benchmark using the commonly accepted maximum token length of 4096 for current LLMs. This threshold serves to categorize graph sizes into \textbf{Within Limit Graph} (\textbf{WL-Graph}) and \textbf{Exceeds Limit Graph} (\textbf{EL-Graph}). WL-Graph ensures that the entire graph can be directly input into LLMs in textual form. Meanwhile, EL-Graph accommodates larger graph structures and we store the graph in files with file paths provided to LLMs. 

Consequently, we have elaborately crafted two types of Graph-Instruction. For WL-Graph, we employ a Two-shot Prompt of extracting graph information for both weighted and unweighted graphs in a list format of NetworkX~\cite{hagberg2008exploring}. By utilizing regular expressions to parse the output text that includes the graph structure information, we can reconstruct the graph in tools (the regular expressions can be found in \textbf{Appendix~\ref{appendixc regular expressions}}). Given the size constraints of EL-Graph, it becomes hard to extract complete graph structure information based on natural language. Therefore, we replace the graph with the file path. For EL-Graph, a one-shot prompt is used to direct LLMs to identify and extract the file path, enabling tools to retrieve graph structure information from the specified path. Two examples for both EL-Graph and WL-Graph can be found in \textbf{Appendix~\ref{appendx Example}}, Table~\ref{Fig: Graph-I-EL}, ~\ref{Fig: Graph-I-WL}.

\textbf{Task-Instruction (for GP)}: To construct the Task-Instruction, we manually create a tool set. For each tool, we define four attributes: Tool Name, Tool Description, Tool Parameters and Return type. This set is intended to inform LLMs about the appropriate graph reasoning tasks for each tool. Based on the predefined tool set, we add some general descriptions of the expected format to constrain the output from LLMs. An example of the set of tools and general descriptions is presented in \textbf{Appendix~\ref{appendx Example}}, Table~\ref{Fig: Task-I-WL}.

\textbf{Parameter-Instruction (for GP)}: For PGQ-Tasks, we specifically employ Parameter-Instruction to further standardize the format of parameters extracted by Task-Instruction. At first, We propose a \textit{Tool Template Retriever}, which identifies the tool name based on previous Task-Instruction and then retrieves the corresponding tool template from the tool set.
Second, we combine the searched tool template with Parameter-Instruction as a new input to get highly accurate tool parameters. An example of the Parameter-Instruction is presented in \textbf{Appendix~\ref{appendx Example}}, Table~\ref{Fig: Para-I-WL}.

\begin{figure}[t]
\centering
\includegraphics[width=3in]{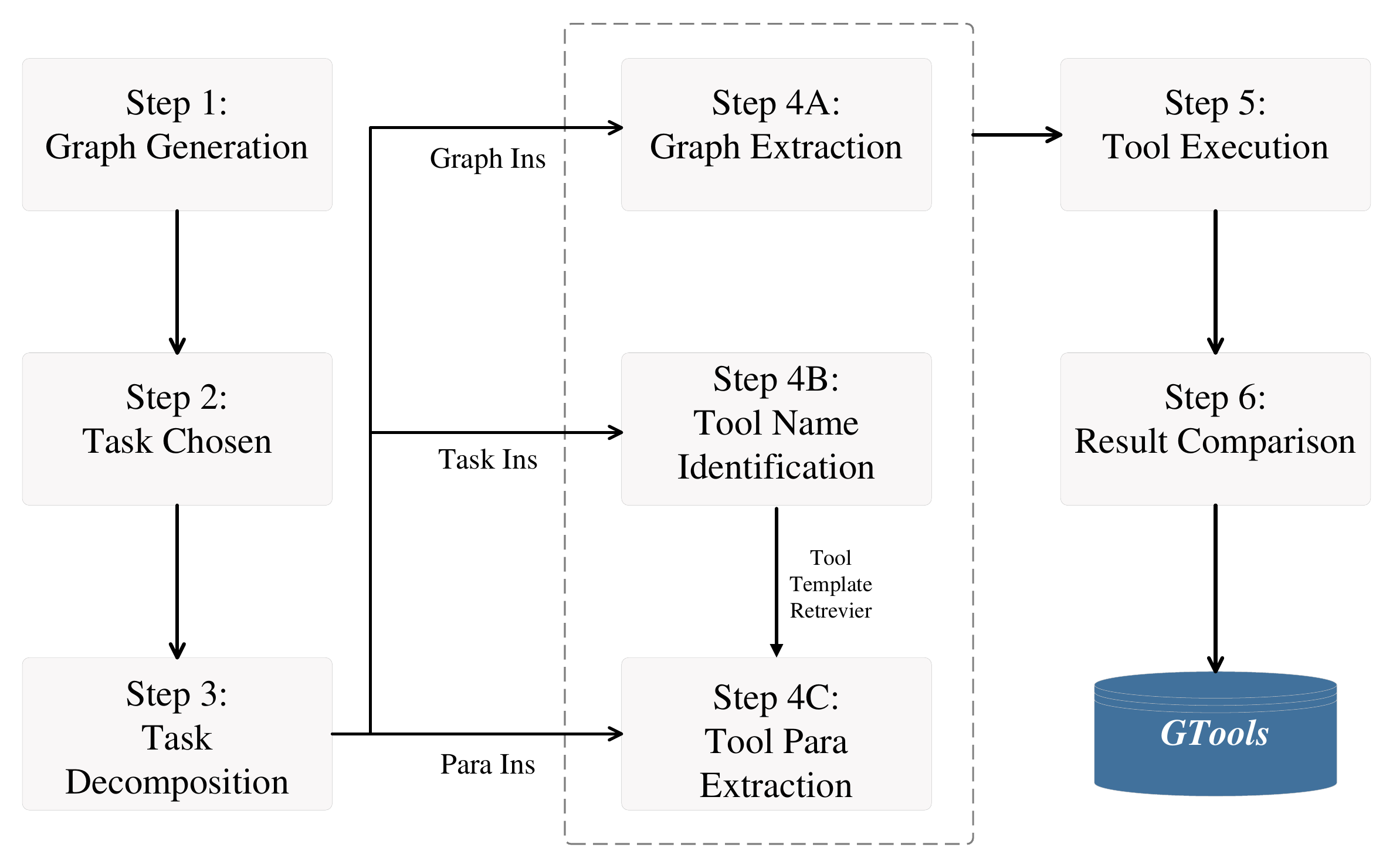}
\caption{Graph reasoning based on GraphTool-Instruction.}  
\vspace{-1.0em}
\Description{Graph reasoning based on GraphTool-Instruction.}  
\label{Fig: workflow}
\end{figure}

\subsection{GTools Construction}
\textbf{Graph Generation}: The graph in tasks is generated based on the number of nodes and the probability of connections between two nodes. Furthermore, we develop specific generation rules tailored to ensure a reasonable distribution of both graph and corresponding answers: 
\begin{itemize}[leftmargin=*]
\item \textbf{Diverse sizes}: For WL-Graph, we ensure that graphs cover a node range from 2 to 40 and an edge range for a maximum of 300 to examine the LLMs' capabilities to handle graph information. Additionally, EL-Graph accommodates larger structures, with node counts ranging from 41 to 100 and up to 1,000 edges.
\item \textbf{Various descriptions}: For each graph reasoning task, we have prepared five different descriptions to evaluate the task identification capabilities of LLMs.
\item \textbf{Balanced answers}: For graph reasoning tasks that determine truth or falsehood, we ensure an even distribution of answers during the graph generation process. This approach aims to prevent LLMs from developing a bias toward any particular type of result, thereby avoiding artificially high accuracy rates.
\item \textbf{Unique answer}: For Topological Sorting, which may have multiple valid solutions, we ensure the uniqueness of the answers during the graph generation process. This facilitates the comparison between the LLMs' results and standard labels. 
% The generation rules for \textit{Topological Sorting} are placed in Appendix~\ref{appendixb Rules of graph generation}.
\end{itemize}

\textbf{Graph Reasoning}: Since the graph and task are determined, an accurate answer label can be obtained through the algorithm program. We use Llama3-8B~\cite{DBLP:journals/corr/abs-2302-13971} as our base model to solve various graph reasoning tasks based on GraphTool-Instruction. The explicit steps of reasoning are presented in Figure~\ref{Fig: workflow}. For each task, we define three subtasks: \textit{graph extraction} $(\mathcal{G})$, \textit{tool name identification} $(\mathcal{N})$ and \textit{tool parameter extraction} $(\mathcal{P})$ represented by a set $S = \{\mathcal{G}, \mathcal{N}, \mathcal{P}\}$. Each subtask's output $g^{(s)}$ is generated by LLMs according to the corresponding instruction $I^{(s)}$ and task $x^{(s)}$:
\begin{align}
g^{(s)} = LLM(I^{(s)},x^{(s)}), s \in \{\mathcal{G}, \mathcal{N}, \mathcal{P}\}.
\end{align}
To transform the text output $g^{(s)}$ into a format available for tools $\bar{g}^{(s)}$, we process the output using regular expressions $R^{(s)}$:
\begin{align}
{\bar{g}^{(s)}}= R^{(s)}(g^{(s)}), s \in \{\mathcal{G}, \mathcal{N}, \mathcal{P}\}.
\end{align}
This allows us to obtain the graph $\bar{g}^{(\mathcal{G})}$, tool name $\bar{g}^{(\mathcal{N})}$ and tool parameters $\bar{g}^{(\mathcal{P})}$. Next, we obtain the output label $\hat{g}$ generated by corresponding tools:
\begin{align}
{\hat{g}}= Tool(\bar{g}^{(\mathcal{G})},\bar{g}^{(\mathcal{N})},\bar{g}^{(\mathcal{P})}).
\end{align}

\textbf{GTools Construction}: To ensure the high quality and reliability of our dataset, we introduce a strict condition for retaining the output $g^{(s)}$. We define a matching function $M$ that checks whether all relevant labels match:
\begin{align}
M(\bar{g}^{(\mathcal{G})}, \bar{g}^{(\mathcal{N})}, \bar{g}^{(\mathcal{P})}, \hat{g}) = \bigwedge_{s \in S} (\bar{g}^{(s)} = g^{*(s)}) \land (\hat{g} = g^*),
\end{align}
where $g^{*(s)}$ refers to the standard labels for each subtask and $g^*$ is the answer label. We define the raw dataset based on graph reasoning tasks $x^{(s)}$ with corresponding instruction $I^{(s)}$ and the output $g^{(s)}$ from LLM as $\mathcal{D}_{raw}$, the definition of $\mathcal{D}_{raw}$ is:
\begin{align}
\mathcal{D}_{\text{raw}} = \left\{ \left\{ (I^{(s)}, x_i^{(s)}, g_i^{(s)}) \mid s \in S \right\} \mid i \in \{1, 2, \ldots, |\mathcal{D}_{\text{raw}}|\} \right\}.
\end{align}
Based on the matching function $M$, we determine whether to retain the output $g^{(s)}$:
\begin{align}
y^{(s)} = 
\begin{cases} 
g^{(s)} & \text{if } M(\bar{g}^{(\mathcal{G})}, \bar{g}^{(\mathcal{N})}, \bar{g}^{(\mathcal{P})}, \hat{g}), \\
\text{discarded} & \text{otherwise}.
\end{cases}
\end{align}
Through the filtering process described above, we obtain the dataset $\mathcal{D}$:
\begin{align}
\mathcal{D} = \bigg\{ &\mathcal{D}_i = \big\{ (I^{(s)}, x_i^{(s)}, y_i^{(s)}) \mid s \in S, \nonumber \\
&y_i^{(s)} = g_i^{(s)} \text{ if } M(\bar{g}^{(\mathcal{G})}, \bar{g}^{(\mathcal{N})}, \bar{g}^{(\mathcal{P})}, \hat{g}) \big\} \mid \mathcal{D}_i \in \mathcal{D}_{\text{raw}} \bigg\}.
\end{align}

After the comparison, we organize the dataset $\mathcal{D}$ in the Alpaca format~\cite{alpaca}, which is commonly used to fine-tune Llama for improving instruction-following capabilities. Finally, we build our dataset, GTools, comprising 40,000 instances, with each task containing 2,000 instances.

\subsection{Fine-tune GraphForge}
 We develop GraphForge by fine-tuning on the GTools dataset using Llama3-8B as the base model. The fine-tuning process employs Low-Rank Adaptation (LoRA) ~\cite{DBLP:conf/iclr/HuSWALWWC22}, a parameter-efficient fine-tuning method that is effective for LLMs. The GTools used for training consist of triples as follows:
\begin{align}
\mathcal{D} = \left\{ \left\{ (I^{(s)}, x_i^{(s)}, y_i^{(s)}) \mid s \in S \right\} \mid i \in \{1, 2, \ldots, |\mathcal{D}|\} \right\},
\end{align}
where $I^{(s)}$ is the instruction for subtask $s$, $x_i^{(s)}$ is the graph reasoning task, and $y_i^{(s)}$ is the expected output for the input $x_i^{(s)}$ under the instruction $I^{(s)}$. 

Formally, for a linear layer represented by $h_{i}^{(s)} = \mathbf{W} \tilde{x}_{i}^{(s)}$ where $\tilde{x}_{i}^{(s)}$ is a combination of the graph reasoning task $x_{i}^{(s)}$ and the instruction $I^{(s)}$, $\mathbf{W} \in \mathbb{R}^{d_{m} \times d_{n}}$ is the weight matrix. LoRA introduces a low-rank update as follows:
\begin{align}
h_{i}^{(s)} = \mathbf{W} \tilde{x}_{i}^{(s)} + \Delta \mathbf{W} \tilde{x}_{i}^{(s)} = \mathbf{W} \tilde{x}_{i}^{(s)} + \frac{\alpha}{r} B A \tilde{x}_{i}^{(s)},
\end{align}
where $h_{i}^{(s)}$ is the actual output by LLMs, $A \in \mathbb{R}^{r \times d_{n}}$ and $B \in \mathbb{R}^{d_{m} \times r}$ are the low-rank matrices, $r$ is the chosen rank significantly smaller than $\min \left(d_{m}, d_{n}\right)$, and $\alpha$ controls the magnitude of the updates to the original weight matrix $\mathbf{W}$. During the learning process, only the matrices $A$ and $B$ are updated.

The cross-entropy loss function measures the discrepancy between the actual labels and the predicted labels. By comparing the LLMs' output $h_{i}^{(s)}$ with the expected output $y_{i}^{(s)}$, we define the loss function for fine-tuning the GraphForge as:
\begin{align}
\mathcal{L}(\mathcal{D})= \frac{1}{|\mathcal{D}|} \sum_{(I^{(s)},x_{i}^{(s)},y_{i}^{(s)})\in \mathcal{D}}\text{cross-entropy}(h_{i}^{(s)},y_{i}^{(s)}), 
\end{align}
\begin{align}
\text{cross-entropy}(h_{i}^{(s)}, y_{i}^{(s)}) = -\sum_{j=1}^{d_m} y_{i}^{(s)}(j) \log h_{i}^{(s)}(j).
\end{align}

During the training process, a learning rate of 1e-5 and a warmup ratio of 0.1 are utilized. The batch size is set to 4, and a cosine learning rate scheduler is implemented to adjust the learning rate cyclically. The GraphForge is fine-tuned over 3 epochs on an 80GB NVIDIA A800 GPU.

\section{Experiments}
\subsection{Experimental Setting}
We conduct extensive experiments to evaluate our method and model, \textit{i.e.}, GraphTool-Instruction and GraphForge, covering both in-domain and out-of-domain tasks. We implement GraphTool-Instruction on three open-source LLMs to validate the effectiveness of our method. Our experiment is designed to answer the following three research questions:
\begin{itemize}[leftmargin=*]
\item \textbf{RQ1 (Main results)}: How does GraphTool-Instruction perform compared to Text-Instruction and Tool-Instruction methods?
\item \textbf{RQ2 (Ablation Study)}: How does GraphTool-Instruction address two challenges (\textit{i.e.}, \textbf{Graph Understanding (GU)} and \textbf{Graph Processing (GP)}) mentioned in section~\ref{Introduction}?
\item \textbf{RQ3 (Error Analysis)}: What are the reasons for the errors in the tool's execution results?
\end{itemize}

\begin{table}[h]
\small
\caption{Baseline methods and corresponding models.}
\begin{tabular}{cccc}
\hline
LLM Type & Method & Base Model & Type \\ \hline
\multirow{2}{*}{Closed-source} & Two-shot & Claude-3-series~\cite{TheC3} & WL \\
 & Two-shot & GPT-series~\cite{DBLP:journals/corr/abs-2303-08774} & WL \\ \cline{2-4} 
\multirow{2}{*}{Text} & NLGraph & GPT-4-turbo~\cite{DBLP:conf/nips/WangFHTHT23} & WL \\
 & GraphWiz & Llama2-13B~\cite{DBLP:journals/corr/abs-2402-16029} & WL \\ \cline{2-4} 
\multirow{4}{*}{Tool} & Function Calling & GPT-3.5-turbo~\cite{DBLP:journals/corr/abs-2303-08774} & WL/EL \\
 & Function Calling & GPT-4o~\cite{DBLP:journals/corr/abs-2303-08774} & WL/EL \\
 & Function Calling & GLM4-0520~\cite{glm2024chatglm} & WL/EL \\
 & Graph-TF & Llama3-8B~\cite{DBLP:journals/corr/abs-2304-11116} & WL/EL \\ \cline{2-4} 
\multirow{3}{*}{GraphTool} & GraphTool & Llama3-8B~\cite{DBLP:journals/corr/abs-2302-13971} & WL/EL \\
 & GraphTool & Llama3.1-8B~\cite{DBLP:journals/corr/abs-2302-13971} & WL/EL \\
 & GraphTool & GLM4-9B~\cite{glm2024chatglm} & WL/EL \\ \hline
\end{tabular}
\label{Table3 Baseline methods and corresponding models.}
\vspace{-2.0em}
\end{table}

\begin{table*}[h]
\small
\caption{Experiment with CoT and Text-Instruction methods on WL-Graph. For conciseness, \textit{Claude-3-haiku}, \textit{Claude-3-sonnet}, \textit{Claude-3-opus}, \textit{GPT-3.5-turbo}, \textit{GPT-4-turbo} and \textit{GraphForge} are abbreviated as Claude-H, Claude-S, Claude-O, GPT-3.5, GPT-4 and GF, respectively. Since we define Maximum Triangle Sum as finding the triangle with the largest sum of edge weights, it is an OOD task for the two Text-Instruction methods, so we have not conducted experiments. The best results of Text-Instruction methods and our method are colored: {\color[HTML]{8E1212}Text},  {\color[HTML]{009901}Ours}.}
\begin{tabular}{cccccccccccc}
\hline
\multicolumn{3}{c}{Task} & \multicolumn{1}{l}{Claude-H} & \multicolumn{1}{l}{Claude-S} & \multicolumn{1}{l}{Claude-O} & \multicolumn{1}{l}{GPT-3.5} & \multicolumn{1}{l}{GPT-4} & \multicolumn{1}{l}{NLGraph} & GraphWiz & $\text{Llama}_{GI}$ & GF \\ \hline
 &  & Cycle & 48.8 & 52.6 & 75.8 & 59.5 & 68.2 & 72.2 & {\color[HTML]{8E1212} 79.8} & 97.8 & {\color[HTML]{009901} 99.0} \\
 & \multirow{-2}{*}{BGA} & Triangle & 16.6 & 13.6 & \color[HTML]{8E1212}21.0 & 10.4 & 20.6 & / & / & 87.6 & {\color[HTML]{009901} 97.8} \\ \cline{2-12} 
 &  & Path & 76.6 & 74.2 & 79.8 & 72.2 & 77.2 & 79.2 & {\color[HTML]{8E1212} 81.0} & 94.6 & {\color[HTML]{009901} 98.8} \\
 &  & Flow & 3.6 & 7.8 & 9.8 & 0.6 & 7.8 & {\color[HTML]{8E1212} 10.8} & 1.0 & 93.4 & {\color[HTML]{009901} 98.6} \\
\multirow{-5}{*}{\rotatebox{90}{Undirected}} & \multirow{-3}{*}{PGQ} & Shortest & 8.8 & 11.4 & 21.2 & 24.4 & 25.8 & {\color[HTML]{8E1212} 28.0} & 7.6 & 93.2 & {\color[HTML]{009901} 98.2} \\ \cline{2-12} 
 &  & Cycle & 50.2 & 53.0 & 73.2 & 68.0 & 64.2 & 70.2 & {\color[HTML]{8E1212} 76.4} & 96.0 & {\color[HTML]{009901} 99.6} \\
 & \multirow{-2}{*}{BGA} & Topo & 16.6 & 17.0 & 22.2 & 24.6 & 34.4 & 34.2 & {\color[HTML]{8E1212} 40.2} & {\color[HTML]{009901} 97.8} & {\color[HTML]{000000} 97.2} \\ \cline{2-12} 
 &  & Path & 73.2 & 77.0 & 83.2 & 67.6 & 84.0 & {\color[HTML]{8E1212} 84.2} & 79.6 & 93.0 & {\color[HTML]{009901} 98.2} \\
 &  & Flow & 4.2 & 7.0 & {\color[HTML]{8E1212} 12.2} & 8.8 & 9.0 & 11.0 & 2.2 & 92.2 & {\color[HTML]{009901} 98.0} \\
\multirow{-5}{*}{\rotatebox{90}{Directed}} & \multirow{-3}{*}{PGQ} & Shortest & 9.6 & 13.8 & 19.0 & 21.2 & 22.6 & {\color[HTML]{8E1212} 26.2} & 6.0 & 93.6 & {\color[HTML]{009901} 98.0} \\ \cline{2-12} 
 & \multicolumn{1}{l}{} & Overall & 30.8 & 32.7 & 41.7 & 35.7 & 41.4 & {\color[HTML]{8E1212} 46.2} & 41.5 & 94.0 & {\color[HTML]{009901} 98.4} \\ \hline
\end{tabular}
\label{Table4 Experiment on CoT methods}
\end{table*}

\begin{table*}[h]
\small
\caption{Experiment with Tool-Instruction and GraphTool-Instruction methods on both WL-Graph and EL-Graph. Symbol * represents closed-source LLMs based on Function Calling (\textit{e.g., GPT-4o* represents GPT-4o-FC}). For conciseness, \textit{GLM4-9B, Llama3-8B, Llama3.1-8B using GraphTool-Instruction without fine-tuning}, Graph-Toolformer and \textit{GraphForge} are abbreviated as $\text{GLM4}_{GI}$, $\text{Llama3}_{GI}$, $\text{Llama3.1}_{GI}$, Graph-TF and GF, respectively. The best results of Tool-Instruction methods and our method in WL-Graph are colored: {\color[HTML]{8E1212}Tool},  {\color[HTML]{009901}Ours} and for EL-Graph: {\color[HTML]{F56B00}Tool},  {\color[HTML]{6200C9}Ours}.}
\begin{tabular}{ccccccccccccccccccc}
\hline
\multicolumn{3}{c}{} & \multicolumn{2}{c}{Graph-TF} & \multicolumn{2}{c}{GLM4-0520*} & \multicolumn{2}{l}{GPT-3.5*} & \multicolumn{2}{c}{GPT-4o*} & \multicolumn{2}{c}{$\text{GLM4}_{GI}$} & \multicolumn{2}{c}{$\text{Llama3}_{GI}$} & \multicolumn{2}{c}{$\text{Llama3.1}_{GI}$} & \multicolumn{2}{c}{GF} \\ \cline{4-19} 
\multicolumn{3}{c}{\multirow{-2}{*}{Task}} & WL & EL & WL & EL & WL & EL & WL & EL & WL & EL & WL & EL & WL & EL & WL & EL \\ \hline
 &  & Cycle & 66.8 & 67.2 & 81.2 & 98.8 & 68.0 & 99.0 & {\color[HTML]{9A0000} 98.8} & {\color[HTML]{F56B00} 99.6} & {\color[HTML]{000000} 99.6} & {\color[HTML]{6200C9} 99.6} & 97.8 & 98.0 & {\color[HTML]{009901} 100} & 99.0 & {\color[HTML]{000000} 99.0} & {\color[HTML]{000000} 99.2} \\
 & \multirow{-2}{*}{BGA} & Triangle & 71.0 & 71.2 & 41.6 & 97.6 & 66.2 & 96.0 & {\color[HTML]{9A0000} 98.8} & {\color[HTML]{F56B00} 100} & 80.2 & 85.2 & 87.6 & 91.4 & 40.2 & 46.2 & {\color[HTML]{009901} 97.8} & {\color[HTML]{6200C9} 99.6} \\ \cline{2-19} 
 &  & Path & 44.2 & 41.6 & 84.6 & {\color[HTML]{F56B00} 99.8} & 59.8 & {\color[HTML]{000000} 99.4} & {\color[HTML]{9A0000} 100} & {\color[HTML]{000000} 99.2} & 86.0 & 88.8 & 94.6 & 97.4 & 96.8 & {\color[HTML]{6200C9} 99.2} & {\color[HTML]{009901} 98.8} & {\color[HTML]{000000} 98.4} \\
 &  & Flow & 38.8 & 40.2 & 37.4 & 96.0 & 52.2 & 96.8 & {\color[HTML]{9A0000} 98.6} & {\color[HTML]{F56B00} 98.8} & 96.2 & {\color[HTML]{6200C9} 100} & 93.4 & 94.2 & 88.8 & 90.0 & {\color[HTML]{009901} 98.6} & {\color[HTML]{000000} 99.2} \\
\multirow{-5}{*}{\rotatebox{90}{Undirected}} & \multirow{-3}{*}{PGQ} & Shortest & 63.8 & 67.0 & 42.8 & 97.6 & 60.8 & 99.8 & {\color[HTML]{9A0000} 98.2} & {\color[HTML]{F56B00} 100} & 96.8 & {\color[HTML]{6200C9} 99.8} & 93.2 & 96.8 & 92.8 & 91.2 & {\color[HTML]{009901} 98.2} & {\color[HTML]{000000} 99.2} \\ \cline{2-19} 
 &  & Cycle & 72.0 & 74.0 & 79.0 & 99.4 & 76.2 & 99.6 & {\color[HTML]{9A0000} 99.4} & {\color[HTML]{F56B00} 100} & 99.6 & 99.8 & 96.0 & 100 & 99.6 & 100 & {\color[HTML]{009901} 99.6} & {\color[HTML]{6200C9} 100} \\
 & \multirow{-2}{*}{BGA} & Topo & 67.0 & 70.2 & 80.0 & {\color[HTML]{F56B00} 99.2} & 69.6 & 98.6 & {\color[HTML]{9A0000} 98.4} & {\color[HTML]{000000} 98.0} & 96.6 & 100 & {\color[HTML]{009901} 97.8} & 99.2 & 96.4 & 95.2 & {\color[HTML]{000000} 97.2} & {\color[HTML]{6200C9} 100} \\ \cline{2-19} 
 &  & Path & 44.8 & 42.6 & 84.4 & 99.6 & 58.4 & 98.4 & {\color[HTML]{9A0000} 100} & {\color[HTML]{F56B00} 100} & 85.2 & 87.6 & 93.0 & 96.6 & 96.2 & 94.4 & {\color[HTML]{009901} 98.2} & {\color[HTML]{6200C9} 99.2} \\
 &  & Flow & 42.6 & 46.2 & 35.8 & 98.6 & 49.4 & 96.6 & {\color[HTML]{9A0000} 98.6} & {\color[HTML]{F56B00} 100} & 95.6 & {\color[HTML]{6200C9} 100} & 92.2 & 94.2 & 90.6 & 92.2 & {\color[HTML]{009901} 98.0} & {\color[HTML]{000000} 97.2} \\
\multirow{-5}{*}{\rotatebox{90}{Directed}} & \multirow{-3}{*}{PGQ} & Shortest & 62.6 & 68.2 & 40.2 & 98.0 & 61.8 & 99.2 & {\color[HTML]{9A0000} 97.2} & {\color[HTML]{F56B00} 99.2} & 96.2 & {\color[HTML]{6200C9} 100} & 93.6 & 96.6 & 93.6 & 94.4 & {\color[HTML]{009901} 98.0} & {\color[HTML]{000000} 98.0} \\ \cline{2-19} 
 & \multicolumn{1}{l}{} & \multicolumn{1}{l}{Overall} & 57.4 & 58.8 & 60.7 & 98.5 & 62.2 & 98.3 & {\color[HTML]{9A0000} 98.8} & {\color[HTML]{F56B00} 99.5} & 93.2 & 96.1 & 94.0 & 96.4 & 89.5 & 90.2 & {\color[HTML]{009901} 98.4} & {\color[HTML]{6200C9} 99.0} \\ \hline
\end{tabular}
\label{Table5 Experiment on Tool-Instruction}
\vspace{-1.0em}
\end{table*}

\textbf{Test Datasets}: According to the existing generation rules of GTools, we have created an additional 500 test instances for each task as our test dataset. Besides, we select Path Existence, Cycle Detection, Topological Sorting, Maximum Flow, and Shortest Path tasks from NLGraph. Since GraphForge has not been trained on NLGraph, we directly combine the test dataset and training dataset of NLGraph for testing. In addition, we select two tasks from NLGraph: Bipartite Graph Matching and GNN, as out-of-domain tasks to validate the effectiveness of GraphForge. (The GNN task in NLGraph is defined as: Given an undirected graph  and a two-dimension node embedding for each node, update the node embedding with the sum of all the neighbors' embeddings.)

\textbf{Baselines}: We carefully select baseline models from four categories for a comprehensive evaluation. We summarize the methods and their base models in Table~\ref{Table3 Baseline methods and corresponding models.}:
\begin{itemize}[leftmargin=*]
\item \textbf{Closed-source LLM}: We selected the most powerful closed-source LLMs currently available. In our experiments, we use the CoT method with a Two-shot prompt based on two series of closed-source LLMs: Claude and GPT. The specific versions of these LLMs are detailed in \textbf{Appendix~\ref{appendx E Details of the closed-source LLM version}}.
\item \textbf{Text-Instruction LLM}: We have also implemented two recently released Text-Instruction methods, NLGraph and GraphWiz with GPT-4-turbo and Llama2-13B. Due to the inherent limitations of these LLMs in text-based reasoning, our experiments are exclusively conducted on WL-Graph.
\item \textbf{Tool-Instruction LLM}: In light of recent support for Function Calling officially released by some closed-source LLMs, we choose GLM4-0520, GPT-3.5-turbo, and GPT-4o as exemplars of the most robust Tool-Instruction LLMs. Additionally, we employ the Graph-ToolFormer. Due to Graph-ToolFormer has only released a model based on GPT-J-6B, we utilize the method of Graph-ToolFormer to fine-tune Llama3-8B on our dataset to ensure a fair comparison of experiment results. For all the aforementioned LLMs, experiments will be conducted on both WL-Graph and EL-Graph. Given the Graph-ToolFormer's deficiency in graph structure extraction capabilities, we employ GraphForge to extract graph structure information for Graph-ToolFormer on WL-Graph. 
\item \textbf{GraphTool-Instruction LLM} To verify the effectiveness of GraphTool-Instruction, we test its performance on GLM4-9B, Llama3-8B and Llama3.1-8B.
\end{itemize}

\subsection{Experimental Setup}
We use an NVIDIA A800 GPU to fine-tune GraphForge and Graph-Toolformer based on LoRA. For inference with all open-source LLMs, we employ a total of 16 NVIDIA Tesla T4 GPUs. For all closed-source LLMs, we leverage the official API interfaces. During inference process, we set the max number of new tokens to 4096, with a sampling parameter top\_p of 1 and a temperature of 0.7.

\subsection{MainResults (RQ1)}
In this experiment, we first evaluate GraphForge compared to state-of-the-art baselines based on Text-Instruction. Due to LLMs' capabilities constraints, we conduct experiments solely on WL-Graph for all LLMs listed in Table~\ref{Table4 Experiment on CoT methods}. Subsequently, we evaluate GraphForge against Tool-Instruction methods. For all LLMs listed in Table~\ref{Table5 Experiment on Tool-Instruction}, experiments are conducted on both WL-Graph and EL-Graph. We employ answer accuracy as the evaluation metric, which calculates the percentage of questions that the LLM correctly predicts out of the total questions in the test dataset. We demonstrate ten types of graph reasoning tasks, and the complete experiment results are in \textbf{Appendix~\ref{appendixa Comprehensive experiment}}. We observe the following results:
\begin{itemize}[leftmargin=*]
\item GraphForge demonstrates graph reasoning capabilities that significantly surpass those of all Text-Instruction methods across various tasks, achieving an average performance of 98.4\%, in contrast to the highest average of 46.2\% recorded by other methods.
\item Compared with Graph-ToolFormer, GraphForge has consistently surpassed by over 40\% on both WL-Graph and EL-Graph. This marked improvement is primarily due to the inherent limitations of the Graph-ToolFormer's method in accurately identifying graph reasoning tasks and extracting tool parameters.
\item In comparison to GPT-4o-FC, GraphForge performs comparably to the high-cost GPT-4o-FC on both WL-Graph and EL-Graph. Compared with GLM4-0520-FC and GPT-3.5-turbo-FC, GraphForge achieves a lead of over 30\% on WL-Graph. It is important to note that GLM4-0520-FC demonstrates poor performance on WL-Graph, with remarkably low accuracies observed in the Maximum Triangle Sum, Maximum Flow and Shortest Path. The same issue has also occurred with GPT-3.5-turbo-FC, where we observe the absence of graph structure information and the inability to parse tool parameters. We have detailed the error analysis in Section \ref{Chapter 4.5Error Analysis} and the case study is in \textbf{Appendix~\ref{appendxf Case study on error}}.
\item Among GLM4-9B, Llama3-8B and Llama3.1-8B based on GraphTool-Instruction. We observe that all three models demonstrate a high accuracy in most tasks. Moreover, the performance of GLM4-9B, when utilizing our approach, surpasses that of GLM4-0520-FC employing Function Calling, thus validating the efficiency of our method. However, we have found GLM-9B and Llama3.1-8B exhibit lower accuracy on some specific tasks, the analysis is in \textbf{Appendix~\ref{appendixa Comprehensive experiment}}.
\end{itemize}

\begin{table}[h]
\small
\caption{Experiment on NLGraph Dataset. Symbol * represents closed-source LLMs based on Function Calling.}
\begin{tabular}{cccccc}
\hline
Type & Task & GPT-4 & GLM4-0520* & GPT-4o* & GF \\ \hline
\multirow{5}{*}{ID} & Cycle & 66.75 & 100 & 100 & 99.73 \\
 & Path & 84.57 & 99.46 & 99.14 & 99.76 \\ \cline{2-6} 
 & Shortest & 51.51 & 86.84 & 99.73 & 99.47 \\
 & Topo & 23.25 & 98.14 & 99.38 & 99.38 \\
 & Flow & 6.57 & 84.86 & 98.28 & 99.14 \\ \cline{2-6} 
\multirow{2}{*}{OOD} & Bipartite & 33.92 & 98.03 & 100 & 99.60 \\
 & GNN & 61.53 & 97.08 & 100 & 99.58 \\ \cline{2-6} 
 & Overall & 46.87 & 94.92 & 99.50 & 99.52 \\ \hline
\end{tabular}
\label{Table6 Experiment On NLGraph dataset.}
\vspace{-1.0em}
\end{table}

Furthermore, we evaluate GraphForge on NLGraph in Table~\ref{Table6 Experiment On NLGraph dataset.}. It is important to note that, for task GNN and Bipartite as out-of-domain tasks, we add the corresponding descriptions of tools into the tool set. GraphForge has demonstrated impressive performance on NLGraph. For OOD tasks, GraphForge also exhibits strong generalization capabilities.

\subsection{Ablation Studies (RQ2)}
In this study, we conduct experiments to assess the effectiveness of Graph-Instruction for challenge \textbf{GU} and Parameter-Instruction for challenge \textbf{GP}. We select two BGA-Tasks and two PGQ-Tasks for detailed analysis of the results in Table~\ref{Table7 Experiment on Ablation Study.}.

% In this study, we conduct experiments to assess the impact of Graph-Instruction and Parameter-Instruction on accuracy improvement. We select two BGA-Tasks and two PGQ-Tasks for detailed analysis of the results in Table~\ref{Table7 Experiment on Ablation Study.}.

\begin{table}[h]
\small
\caption{Experiment on Ablation Study. GI refers to Graph-Instruction and PI refers to Parameter-Instruction.}
\begin{tabular}{ccccc}
\hline
Task/Variants & GF & $w/o$ GI & $w/o$ PI & $w/o$ GI, PI \\ \hline
Cycle & 99.0 & 86.2 & / & 86.2 \\
Triangle & 97.8 & 66.8 & / & 66.8 \\
Path & 98.8 & 52.0 & 89.0 & 51.2 \\
Shortest & 98.2 & 40.6 & 92.2 & 33.6 \\ \hline
\end{tabular}
\label{Table7 Experiment on Ablation Study.}
\vspace{-2.0em}
\end{table}

We observe a substantial enhancement with Graph-Instruction which demonstrates its effectiveness in facing \textbf{GU} challenge. Upon examining the outputs, we find that the absence of Graph-Instruction leads to unparseable issue about graph structure information. This issue arises from the LLMs' inability to concurrently and accurately execute three subtasks: \textit{graph extraction}, \textit{tool name identification} and \textit{tool parameter extraction}. The implementation of Parameter-Instruction also results in about 8\% improvement in accuracy which demonstrates its effectiveness in facing \textbf{GP} challenge. This enhancement is because LLMs still cannot fully adhere to the instruction, often encountering issues in parameters extracting such as misalignment, omissions, and inconsistency with predefined formats.

\begin{figure}[h]
\centering 
\includegraphics[width=3.0in]{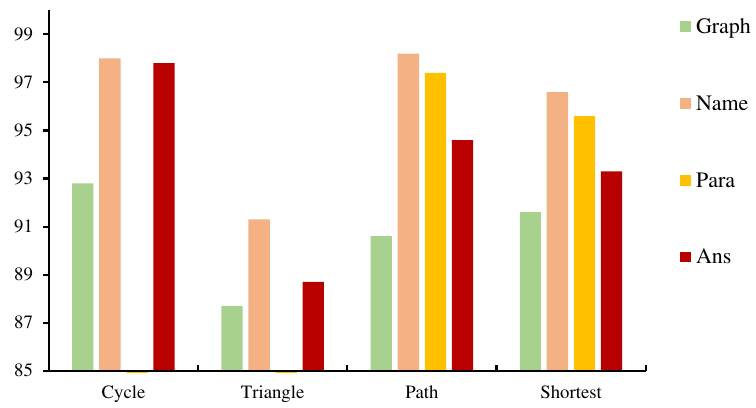}%%%%%%%%%%%%%%%%scale=缩小比例，或者用width=2in
\vspace{-1.0em}
\caption{Impact of graph, name and parameter accuracies on overall answer accuracy. Notably, both Cycle Detection and Maximum Triangle Sum are BGA-Task, so there is no result for Parameter Accuracy.}  
\vspace{-1.0em}
\label{Fig: accuracydiversity}
\end{figure}

\begin{figure}[t]
\centering 
\includegraphics[width=3.2in]{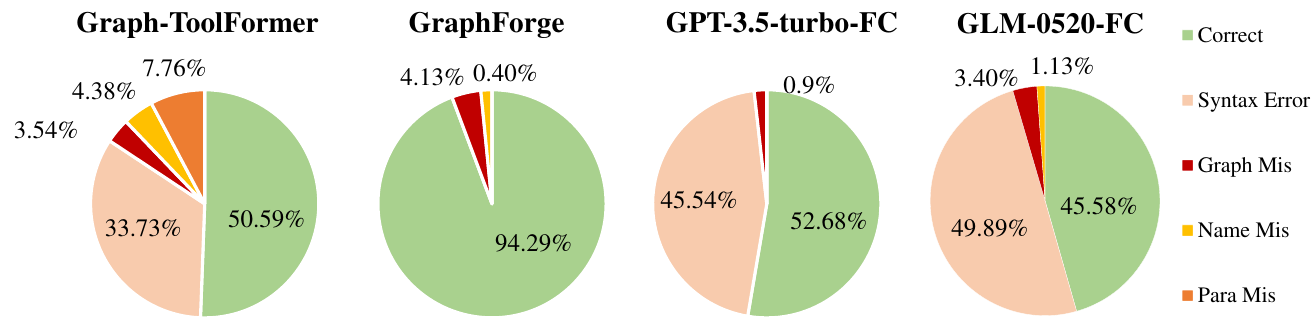}%%%%%%%%%%%%%%%%scale=缩小比例，或者用width=2in
\caption{Error Analysis on GPT-3.5-turbo-FC, GLM-0520-FC, Graph-ToolFormer and GraphForge. Mis is short for Mismacth.}  
\vspace{-1.5em}
\label{Fig: Error Analysis}
\end{figure}

We conduct further analysis to identify factors that influence the accuracy of answers in Figure~\ref{Fig: accuracydiversity}. For this purpose, we introduce three accuracy metrics: \textit{Graph Accuracy}, \textit{Name Accuracy} and \textit{Parameter Accuracy}. Notably, for assessing whether graphs match, we directly compare the list of edges extracted by LLMs with the standard list. We consider the graphs inconsistent if there is any discrepancy in even a single edge. The experiment is conducted on Llama3-8B with GraphTool-Instruction.

We observe that \textit{Name Accuracy} significantly surpasses the other metrics, indicating that LLMs possess robust task identification capabilities. Due to the Parameter-Instruction relying on LLMs' identification of the tool name, \textit{Parameter Accuracy} is slightly lower than \textit{Name Accuracy}. For tasks that rely on the entire information of graph structure such as Maximum Triangle Sum and Shortest Path. The \textit{Answer Accuracy} depends on \textit{Graph Accuracy}. Inaccurate extraction of graph structure information can lead to incorrect reasoning results, even if other parameters are extracted correctly. On the other hand, for tasks like Cycle Detection, which are less dependent on the entire graph structure information, there is a notable disparity where \textit{Answer Accuracy} surpasses \textit{Graph Accuracy}. Thus, improving the precision of graph structure extraction could be crucial for enhancing the overall performance of LLMs.

\subsection{Error Analysis (RQ3)}
\label{Chapter 4.5Error Analysis}
To systematically analyze the shortcomings of Tool-Instruction methods and our method in WL-Graph, we select the Shortest Path as representative. We manually examine the results and classify them into five categories:
\begin{itemize}[leftmargin=*]
\item \textbf{Correct}: When the final result matches the answer, the current result is considered correct.
\item \textbf{Syntax Error}: A Syntax Error is considered to have occurred in the following two scenarios: (1) For Function Calling, if the argument tag returns empty; (2) The predefined regular expression encounters parsing errors for graphs, names or parameters.
\item \textbf{Graph Mismatch}: A Graph Mismatch is considered to have occurred when the graph extracted by LLMs does not match the graph in the task, even if the answer is correct.
\item \textbf{Name Mismatch}: A Name Mismatch is considered to have occurred when the tool name identified by LLMs does not match the given name in the task, even if the answer is correct.
\item \textbf{Para Mismatch}: A Parameter Mismatch is considered to have occurred when there is a mismatch in parameters, even if the answer is correct.
\end{itemize} 

We show the statistics in Figure~\ref{Fig:  Error Analysis} and error cases in \textbf{Appendix~\ref{appendxf Case study on error}}. Among the four methods, the distributions of error types are different:
\begin{itemize}[leftmargin=*]
\item GLM4-0520-FC and GPT-3.5-turbo-FC exhibit similar distributions of error types. However, there are significant differences in the causes of these errors. For GLM4-0520-FC, during the Function Calling API invocation process, even setting the API retry count to 5 to eliminate network-related reasons, there are multiple instances where the arguments return null. We speculate that this is due to the internal parameter parsing mechanism of GLM4-0520 not handling graph structure information with weights specifically, leading to numerous Syntax Errors. While GPT-3.5-turbo-FC behaves differently. Both GPT-3.5-turbo-FC and GPT-4-turbo-FC's output contains significant information gaps and omissions even based on a two-shot prompt for graph structure extraction, leading to various Syntax Errors. However, we have discovered the impressive ability of GPT-4o-FC to follow Tool-Instruction, which achieves remarkable accuracy in various graph reasoning tasks.
\item For Graph-Toolformer, even if we mitigate its shortcomings by providing graph structure information in advance, we observe that Graph-Toolformer has lower performance. After manually checking the output, we find the reason for this result is that Graph-Toolformer often generates misalignment, omissions, and inconsistencies with the predefined tool formats.
\item GraphForge is mainly encountered with Graph Mismatch. This issue tends to occur in scenarios where the number of edges in the graph exceeds one hundred, indicating that the model demonstrates weaker graph structure extraction capabilities when dealing with very long inputs. 
\end{itemize}

\section{Conclusion}
In this work, we propose an innovative GraphTool-Instruction to enhance LLMs' graph reasoning capabilities. Experiment results have demonstrated the robustness of our method, which notably surpasses existing Text-Instruction and Tool-Instruction instruction methods. We have also proposed our dataset GTools, which encompasses twenty graph reasoning tasks, further enhancing the capabilities of LLMs in reasoning on graph tasks. Three new evaluation metrics: \textit{Graph}, \textit{Tool Name} and \textit{Tool Parameter} are employed to ensure the reliability of our dataset. Furthermore, our model GraphForge fine-tuned on the GTools, showcases outstanding performance, achieving an accuracy of more than 98\% accuracy. 
Our future work aims to design a framework that can accommodate a greater variety of graph tools and enhance the capabilities of LLMs to tackle real graph reasoning challenges such as Recommendation and Knowledge Graph.

\section{Acknowledgement}
This work is supported by National Natural Science Foundation of China (No.62406057, No. 62176046), Fundamental Research Funds for the Central Universities No.ZYGX2023K010 and Noncommunicable Chronic Diseases-National Science and Technology Major Project (2023ZD0501806).

\bibliographystyle{ACM-Reference-Format}
\bibliography{acmart}

\appendix

\section{Comprehensive Experiment Results with Baselines}
\label{appendixa Comprehensive experiment}
We conduct comprehensive experiments on all twenty tasks. Due to the weak OOD task performance of NLGraph and GraphWiz, we have not conducted evaluation on the newly added tasks. Considering these results, we have some new observations:

\begin{table*}[t]
\small
\caption{Experiment with CoT and Text-Instruction methods on WL-Graph. The uppercase letter C represents Count in tasks, and the uppercase letter E represents Existence in tasks. For conciseness, \textit{Claude-3-haiku}, \textit{Claude-3-sonnet}, \textit{Claude-3-opus}, \textit{GPT-3.5-turbo}, \textit{GPT-4-turbo} and \textit{GraphForge} are abbreviated as Claude-H, Claude-S, Claude-O, GPT-3.5, GPT-4 and GF, respectively. For the OOD tasks of two Text-Instruction methods, we have not conducted experiments. The best results of Text-Instruction methods and our method are colored: {\color[HTML]{8E1212}Text},  {\color[HTML]{009901}Ours}.}
\begin{tabular}{cccccccccccc}
\hline
\multicolumn{3}{c}{Task} & \multicolumn{1}{l}{Claude-H} & \multicolumn{1}{l}{Claude-S} & \multicolumn{1}{l}{Claude-O} & \multicolumn{1}{l}{GPT3.5} & \multicolumn{1}{l}{GPT-4} & \multicolumn{1}{l}{NLGraph} & GraphWiz & $\text{Llama}_{GI}$ & GF \\ \hline
 &  & Cycle & 48.8 & 52.6 & 75.8 & 59.5 & 68.2 & 72.2 & {\color[HTML]{8E1212} 79.8} & 97.8 & {\color[HTML]{009901} 99.0} \\
 &  & Triangle & 16.6 & 13.6 & {\color[HTML]{8E1212} 21.0} & 10.4 & 20.6 & / & / & 87.6 & {\color[HTML]{009901} 97.8} \\
 &  & Edge C & 5.4 & 18.4 & {\color[HTML]{8E1212} 22.4} & 23.0 & 22.2 & / & / & 98.7 & {\color[HTML]{009901} 98.8} \\
 & \multirow{-4}{*}{BGA} & Node C & 100 & 100 & 100 & 100 & {\color[HTML]{8E1212} 100} & / & / & 98.8 & {\color[HTML]{009901} 100} \\ \cline{2-12} 
 &  & Degree C & 4.2 & 16.6 & {\color[HTML]{8E1212} 32.2} & 8.2 & 29.8 & / & / & 96.4 & {\color[HTML]{009901} 97.8} \\
 &  & Edge E & 76.2 & 83.2 & 94.2 & 82.6 & {\color[HTML]{8E1212} 96.4} & / & / & 90.0 & {\color[HTML]{009901} 94.2} \\
 &  & Node E & 100 & 100 & 100 & 100 & {\color[HTML]{8E1212} 100} & / & / & 98.2 & {\color[HTML]{009901} 100} \\
 &  & Path E & 76.6 & 74.2 & 79.8 & 72.2 & 77.2 & 79.2 & {\color[HTML]{8E1212} 81.0} & 94.6 & {\color[HTML]{009901} 98.8} \\
 &  & Flow & 3.6 & 7.8 & 9.8 & 0.6 & 7.8 & {\color[HTML]{8E1212} 10.8} & 1.0 & 93.4 & {\color[HTML]{009901} 98.6} \\
\multirow{-10}{*}{\rotatebox{90}{Undirected}} & \multirow{-6}{*}{PGQ} & Shortest & 8.8 & 11.4 & 21.2 & 24.4 & 25.8 & {\color[HTML]{8E1212} 28.0} & 7.6 & 93.2 & {\color[HTML]{009901} 98.2} \\ \cline{2-12} 
 &  & Cycle & 50.2 & 53.0 & 73.2 & 68.0 & 64.2 & 70.2 & {\color[HTML]{8E1212} 76.4} & 96.0 & {\color[HTML]{009901} 99.6} \\
 &  & Topo & 16.6 & 17.0 & 22.2 & 24.6 & 34.4 & 34.2 & {\color[HTML]{8E1212} 40.2} & {\color[HTML]{009901} 97.8} & {\color[HTML]{000000} 97.2} \\
 &  & Edge C & 6.0 & 16.2 & {\color[HTML]{8E1212} 23.4} & 21.4 & 23.2 & / & / & 98.0 & {\color[HTML]{009901} 96.6} \\
 & \multirow{-4}{*}{BGA} & Node C & 100 & 100 & 100 & 100 & {\color[HTML]{8E1212} 100} & / & / & 98.8 & {\color[HTML]{009901} 99.4} \\ \cline{2-12} 
 &  & Degree C & 4.0 & 15.4 & 29.4 & 7.2 & {\color[HTML]{8E1212} 32.2} & / & / & 94.4 & {\color[HTML]{009901} 96.6} \\
 &  & Edge E & 78.2 & 85.5 & 93.6 & 82.0 & {\color[HTML]{8E1212} 95.8} & / & / & 92.3 & {\color[HTML]{009901} 97.0} \\
 &  & Node E & 100 & 100 & 100 & 100 & {\color[HTML]{8E1212} 100} & / & / & 99.6 & {\color[HTML]{009901} 100} \\
 &  & Path E & 73.2 & 77.0 & 83.2 & 67.6 & 84.0 & {\color[HTML]{8E1212} 84.2} & 79.6 & 93.0 & {\color[HTML]{009901} 98.2} \\
 &  & Flow & 4.2 & 7.0 & {\color[HTML]{8E1212} 12.2} & 8.8 & 9.0 & 11.0 & 2.2 & 92.2 & {\color[HTML]{009901} 98.0} \\
\multirow{-10}{*}{\rotatebox{90}{Directed}} & \multirow{-6}{*}{PGQ} & Shortest & 9.6 & 13.8 & 19.0 & 21.2 & 22.6 & {\color[HTML]{8E1212} 26.2} & 6.0 & 93.6 & {\color[HTML]{009901} 98.0} \\ \cline{2-12} 
 & \multicolumn{1}{l}{} & Overall & 44.1 & 48.1 & 55.6 & 49.1 & \color[HTML]{8E1212}55.7 & 46.2 & 41.5 & 95.2 & {\color[HTML]{009901} 98.2} \\ \hline
\end{tabular}
\label{Table3 complete}
\end{table*}

\begin{table*}[t]
\small
\caption{Experiment with Tool-Instruction and GraphTool-Instruction methods on both WL-Graph and EL-Graph. Symbol * represents model based on Function Calling. For conciseness, \textit{Llama3-8B using GraphTool-Instruction without fine-tunind} and \textit{GraphForge} are abbreviated as $\text{Llama}_{GI}$ and GF, respectively. The best results of Tool-Instruction methods and our method in WL-Graph are colored: {\color[HTML]{8E1212}Tool},  {\color[HTML]{009901}Ours} and for EL-Graph: {\color[HTML]{F56B00}Tool},  {\color[HTML]{6200C9}Ours}.}
\begin{tabular}{ccccccccccccccccccc}
\hline
\multicolumn{3}{c}{} & \multicolumn{2}{c}{Graph-TF} & \multicolumn{2}{c}{GLM4-0520*} & \multicolumn{2}{c}{GPT3.5*} & \multicolumn{2}{c}{GPT4o*} & \multicolumn{2}{c}{$\text{GLM}_{GI}$} & \multicolumn{2}{c}{$\text{Llama3}_{GI}$} & \multicolumn{2}{c}{$\text{Llama3.1}_{GI}$} & \multicolumn{2}{c}{GF} \\ \cline{4-19} 
\multicolumn{3}{c}{\multirow{-2}{*}{Task}} & WL & EL & WL & EL & WL & EL & WL & EL & WL & EL & WL & EL & WL & EL & WL & EL \\ \hline
 &  & Cycle & 66.8 & 67.2 & 81.2 & 98.8 & 68.0 & 99.0 & {\color[HTML]{8E1212} 98.8} & {\color[HTML]{F56B00} 99.6} & 99.6 & {\color[HTML]{6200C9} 99.6} & 97.8 & 98.0 & {\color[HTML]{009901} 100} & 99.0 & 99.0 & 99.2 \\
 &  & Triangle & 71.0 & 71.2 & 41.6 & 97.6 & 66.2 & 96.0 & {\color[HTML]{8E1212} 98.8} & {\color[HTML]{F56B00} 100} & 80.2 & 85.2 & 87.6 & 91.4 & 40.2 & 46.2 & {\color[HTML]{009901} 97.8} & {\color[HTML]{6200C9} 99.6} \\
 &  & Edge C & 75.6 & 70.4 & 88.0 & 99.0 & 68.8 & 100 & {\color[HTML]{8E1212} 99.2} & {\color[HTML]{F56B00} 100} & 94.6 & 100 & 98.7 & 98.0 & 92.8 & 99.2 & {\color[HTML]{009901} 98.8} & {\color[HTML]{6200C9} 100} \\
 & \multirow{-4}{*}{BGA} & Node C & 77.2 & 78.0 & 90.0 & 100 & 71.6 & 100 & {\color[HTML]{8E1212} 100} & {\color[HTML]{F56B00} 100} & 94.8 & 100 & 98.8 & 97.9 & 98.8 & 99.4 & {\color[HTML]{009901} 100} & {\color[HTML]{6200C9} 100} \\ \cline{2-19} 
 &  & Degree C & 50.0 & 50.6 & 90.0 & 97.9 & 58.0 & 99.2 & {\color[HTML]{8E1212} 99.2} & {\color[HTML]{F56B00} 100} & 92.0 & {\color[HTML]{6200C9} 99.2} & 96.4 & 96.8 & 80.0 & 88.8 & {\color[HTML]{009901} 97.8} & 98.8 \\
 &  & Edge E & \multicolumn{1}{l}{48.0} & \multicolumn{1}{l}{46.8} & 94.0 & 96.0 & \multicolumn{1}{l}{64.8} & 99.4 & {\color[HTML]{8E1212} 100} & {\color[HTML]{F56B00} 100} & 97.2 & 99.2 & 90.0 & 92.2 & {\color[HTML]{009901} 98.2} & 100 & 94.2 & {\color[HTML]{6200C9} 99.2} \\
 &  & Node E & \multicolumn{1}{l}{50.2} & \multicolumn{1}{l}{48.9} & 81.0 & 99.2 & 60.2 & 100 & {\color[HTML]{8E1212} 100} & {\color[HTML]{F56B00} 100} & 97.6 & 100 & 98.2 & 98.8 & 99.8 & 100 & {\color[HTML]{009901} 100} & {\color[HTML]{6200C9} 100} \\
 &  & Path & 44.2 & 41.6 & 84.6 &\color[HTML]{F56B00} 99.8 & 59.8 & {99.4} & {\color[HTML]{8E1212} 100} & 99.2 & 86.0 & 88.8 & 94.6 & 97.4 & 96.8 & {\color[HTML]{6200C9} 99.2} & {\color[HTML]{009901} 98.8} & 98.4 \\
 &  & Flow & 38.8 & 40.2 & 37.4 & 96.0 & 52.2 & 96.8 & {\color[HTML]{8E1212} 98.6} & {\color[HTML]{F56B00} 98.8} & 96.2 & {\color[HTML]{6200C9} 100} & 93.4 & 94.2 & 88.8 & 90.0 & {\color[HTML]{009901} 98.6} & 99.2 \\
\multirow{-10}{*}{\rotatebox{90}{Undirected}} & \multirow{-6}{*}{PGQ} & Shortest & 63.8 & 67.0 & 42.8 & 97.6 & 60.8 & 99.8 & {\color[HTML]{8E1212} 98.2} & {\color[HTML]{F56B00} 100} & 96.8 & {\color[HTML]{6200C9} 99.8} & 93.2 & 96.8 & 92.8 & 91.2 & {\color[HTML]{009901} 98.2} & 99.2 \\ \cline{2-19} 
 &  & Cycle & 72.0 & 74.0 & 79.0 & 99.4 & 76.2 & 99.6 & {\color[HTML]{8E1212} 99.4} & {\color[HTML]{F56B00} 100} & 99.6 & 99.8 & 96.0 & 100 & 99.6 & 100 & {\color[HTML]{009901} 99.6} & {\color[HTML]{6200C9} 100} \\
 &  & Topo & 67.0 & 70.2 & 80.0 & \color[HTML]{F56B00} 99.2 & 69.6 & { 98.6} & {\color[HTML]{8E1212} 98.4} & 98.0 & 96.6 & 100 & {\color[HTML]{009901} 97.8} & 99.2 & 96.4 & 95.2 & 97.2 & {\color[HTML]{6200C9} 100} \\
 &  & Edge C & 70.6 & 74.6 & 79.0 & 98.9 & 64.0 & 100 & {\color[HTML]{8E1212} 98.9} & {\color[HTML]{F56B00} 100} & 94.8 & 100 & {\color[HTML]{009901} 98.0} & 98.9 & 93.2 & 99.2 & 96.6 & {\color[HTML]{6200C9} 100} \\
 & \multirow{-4}{*}{BGA} & Node C & 77.2 & 78.0 & 80.0 & 100 & 70.8 & 100 & {\color[HTML]{8E1212} 99.8} & {\color[HTML]{F56B00} 100} & 94.2 & 100 & 98.8 & 100 & 99.2 & 99.2 & {\color[HTML]{009901} 99.4} & {\color[HTML]{6200C9} 100} \\ \cline{2-19} 
 &  & Degree C & 51.0 & 50.8 & 87.0 & 100 & 58.8 & 98.8 & {\color[HTML]{8E1212} 98.6} & {\color[HTML]{F56B00} 100} & 84.8 & 99.2 & 94.4 & 96.0 & 83.0 & 89.6 & {\color[HTML]{009901} 96.6} & {\color[HTML]{6434FC} 99.4} \\
 &  & Edge E & \multicolumn{1}{l}{46.2} & \multicolumn{1}{l}{46.0} & 81.0 & 94.6 & 62.0 & 99.2 & {\color[HTML]{8E1212} 99.6} & {\color[HTML]{F56B00} 100} & {\color[HTML]{009901} 97.2} & 98.8 & 92.3 & 94.5 & {\color[HTML]{6200C9} 98.8} & 100 & 97.0 & 97.0 \\
 &  & Node E & \multicolumn{1}{l}{50.6} & \multicolumn{1}{l}{52.6} & 82.2 & 99.2 & \multicolumn{1}{l}{64.4} & \multicolumn{1}{l}{100} & {\color[HTML]{8E1212} 99.8} & {\color[HTML]{F56B00} 100} & 97.2 & 100 & 99.6 & 100 & 99.8 & 100 & {\color[HTML]{009901} 100} & {\color[HTML]{6200C9} 100} \\
 &  & Path & 44.8 & 42.6 & 84.4 & 99.6 & 58.4 & 98.4 & {\color[HTML]{8E1212} 100} & {\color[HTML]{F56B00} 100} & 85.2 & 87.6 & 93.0 & 96.6 & 96.2 & 94.4 & {\color[HTML]{009901} 98.2} & {\color[HTML]{6200C9} 99.2} \\
 &  & Flow & 42.6 & 46.2 & 35.8 & 98.6 & 49.4 & 96.6 & {\color[HTML]{8E1212} 98.6} & {\color[HTML]{F56B00} 100} & 95.6 & {\color[HTML]{6200C9} 100} & 92.2 & 94.2 & 90.6 & 92.2 & 98.0 & 97.2 \\
\multirow{-10}{*}{\rotatebox{90}{Directed}} & \multirow{-6}{*}{PGQ} & Shortest & 62.6 & 68.2 & 40.2 & 98.0 & 61.8 & 99.2 & {\color[HTML]{8E1212} 97.2} & {\color[HTML]{F56B00} 99.2} & 96.2 & {\color[HTML]{6200C9} 100} & 93.6 & 96.6 & 93.6 & 94.4 & 98.0 & 98.0 \\ \cline{2-19} 
 & \multicolumn{1}{l}{} & Overall & 58.5 & 59.3 & 73.0 & 98.5 & 63.3 & 99.0 & {\color[HTML]{8E1212} 99.2} & {\color[HTML]{F56B00} 99.7} & 93.8 & 97.9 & 95.2 & 96.7 & 92.0 & 93.9 & {\color[HTML]{009901} 98.2} & {\color[HTML]{6200C9} 99.2} \\ \hline
\end{tabular}
\label{Table4 complete}
% \vspace{-1.0em}
\end{table*}

\begin{itemize}[leftmargin=*]
\item Table~\ref{Table3 complete} shows that Claude and GPT demonstrate extremely strong performance on Node Count and Edge Count tasks. This is because such tasks only require models to focus on very limited but key information. The slightly lagging results from some Tool-Instruction methods are mainly due to incorrect tool name identification, which leads to outcomes that are irrelevant to the tasks.
\item Table~\ref{Table4 complete} shows that GLM-9B and Llama3.1-8B exhibit lower accuracy on tasks such as Maximum Triangle Sum, Degree Count, Shortest Path and Maximum Flow. Upon analysis of experiment results, we discover that these two models tend to decompose complex tasks. For example, in the Shortest Path, the models initially use tools to determine if a path exists between nodes before attempting to compute the shortest path. We consider this to be a rational step in reasoning. However, due to the lack of a multi-step reasoning mechanism, we regard such steps as incorrect during the accuracy evaluation process.
\end{itemize}

\section{Regular Expressions for Information Extraction}
\label{appendixc regular expressions}
We define the Graph extract regular expression as:
\begin{align}
\text{\textdollar(\textbackslash d+), (\textbackslash d+)}.
\end{align}
For tasks with edges that have weights, we add \textit{weight} to the regular expression to extract the starting and ending nodes of the edges as well as the edge weights. Specifically, for the maximum flow problem, we use \textit{capacity }instead of \textit{weight}:
\begin{align}
\text{\textdollar(\textbackslash d+), (\textbackslash d+), \{\textquotesingle weight\textquotesingle:\textbackslash s*(\textbackslash d+)\}},
\end{align}
\begin{align}
\text{\textdollar(\textbackslash d+), (\textbackslash d+), \{\textquotesingle capacity\textquotesingle:\textbackslash s*(\textbackslash d+)\}}.
\end{align}
We define the tool name extraction regular expression as:
\begin{align}
\text{API\_name:\textbackslash s*(\textbackslash w+|\textbackslash n\textbackslash s*\textbackslash w+)"}.
\end{align}
For parameter extraction, we define a general template, the number of parameters to be extracted will be adjusted based on the needs of the task:
\begin{align}
&\text{(?:source\textbackslash s*=\textbackslash s*(\textbackslash d+)[,\textbackslash s]*target\textbackslash s*=\textbackslash s*(\textbackslash d+))}.
\end{align}
Additionally, to address the potential omission of parameter names during the model's output process, we design a general template to enhance the accuracy of parameter extraction.
\begin{align}
&\text{(?:(?:G,\textbackslash s*(\textbackslash d+),\textbackslash s*(\textbackslash d+)}. 
\end{align}

\section{Detailed Graph Reasoning Tasks Definition}
\label{appendixd Details of the graph reasoning task definition}
The details of twenty graph reasoning tasks are listed in Table~\ref{Table10 Details of Graph Reasoning Tasks}.

\begin{table*}[h]
\caption{Details of Graph Reasoning Tasks.}
\begin{tabular}{p{2cm}>{\centering\arraybackslash}p{1cm}>{\centering\arraybackslash}p{1cm}>{\centering\arraybackslash}p{1cm}p{6cm}c>{\centering\arraybackslash}p{2cm}}
\hline
\multirow{2}{*}{Task} & \multirow{2}{*}{Weighted} & \multicolumn{2}{c}{Graph Type} & \multicolumn{1}{c}{\multirow{2}{*}{Description}} & \multicolumn{1}{l}{\multirow{2}{*}{Tool Algorithm}} & \multicolumn{1}{l}{\multirow{2}{*}{Time Complexity}} \\
 &  & Directed & Undirected & \multicolumn{1}{c}{} & \multicolumn{1}{l}{} & \multicolumn{1}{l}{} \\ \hline
Cycle Detection & \XSolidBrush & \Checkmark & \Checkmark & Determine whether there exists any cycle in a given graph. & DFS & $O(|V|+|E|)$ \\
Maximum Triangle Sum & \Checkmark & \XSolidBrush & \Checkmark & Finding the triangle with the largest sum of edge weights in a given graph. & Brute-force Search & $O({|V|}^3)$ \\
Edge Count & \XSolidBrush & \Checkmark & \Checkmark & Count the total number of edges in a given graph. & Direct Lookup & $O(1)$ \\
Node Count & \XSolidBrush & \Checkmark & \Checkmark & Count the total number of nodes in a given graph. & Direct Lookup & $O(1)$ \\
Topological Sorting & \XSolidBrush & \Checkmark & \XSolidBrush & Arrange the nodes of a given graph in the topological order. & DFS & $O(|V|+|E|)$ \\
Degree Count & \XSolidBrush & \Checkmark & \Checkmark & Count the number of edges connected to a specific node in a given graph. & Direct Lookup & $O(1)$ \\
Edge Existence & \XSolidBrush & \Checkmark & \Checkmark & Determine whether a specific edge exists between two nodes in a given graph. & Direct Lookup & $O(1)$ \\
Node Existence & \XSolidBrush & \Checkmark & \Checkmark & Determine whether a specific node exists in a given graph. & Direct Lookup & $O(1)$ \\
Maximum Flow & \Checkmark & \Checkmark & \Checkmark & Determine the largest amount of flow from a source node to a sink node. & Edmonds-Karp & $O(|V| \cdot {|E|}^2)$ \\
Path Existence & \XSolidBrush & \Checkmark & \Checkmark & Determine whether a specific path exists between two nodes in a given graph. & DFS & $O(|V|+|E|)$ \\
Shortest Path & \Checkmark & \Checkmark & \Checkmark & Determine the minimum distance between two nodes in a given graph. & Dijkstra & $O({|V|}^2+|E|)$ \\ \hline
\end{tabular}
\label{Table10 Details of Graph Reasoning Tasks}
% \vspace{-1.0em}
\end{table*}

\section{Detailed Version of LLMs}
\label{appendx E Details of the closed-source LLM version} 
Table~\ref{Table Details of the closed-source LLM version} presents the detailed version of both open-source and closed-source LLMs used in our experiments. 

\begin{table*}[t]
\caption{Baseline methods and corresponding models.}
\begin{tabular}{ccc}
\hline
Method & Base Model & Detail Version \\ \hline
\multirow{5}{*}{Two-shot prompt} & Claude-3-haiku & claude-3-haiku-20240307 \\
 & Claude-3-sonnet & claude-3-sonnet-20240229 \\
 & Claude-3-opus & claude-3-opus-20240229 \\
 & GPT-3.5-turbo & gpt-3.5-turbo-0125 \\
 & GPT-4-turbo & gpt-4-turbo-2024-04-09 \\ \hline
\multirow{3}{*}{Function Calling} & GPT-3.5-turbo-FC & gpt-3.5-turbo-0125 \\
 & GPT-4-turbo-FC & gpt-4-turbo-2024-04-09 \\
 & GLM4-0520-FC & GLM4-0520 \\ \hline
\multirow{3}{*}{GraphTool-Instruction} & GLM4-9B & GLM4-9B-Chat \\
 & Llama3-8B & Llama3-8B-Instruct \\
 & Llama3.1-8B & Llama3.1-8B-Instruct \\ \hline
\end{tabular}
\label{Table Details of the closed-source LLM version}
\vspace{-1.0em}
\end{table*}

\section{Examples of GraphTool-Instruction}
\label{appendx Example}
\begin{figure*}[t]
\centering 
\includegraphics[width=6.5in]{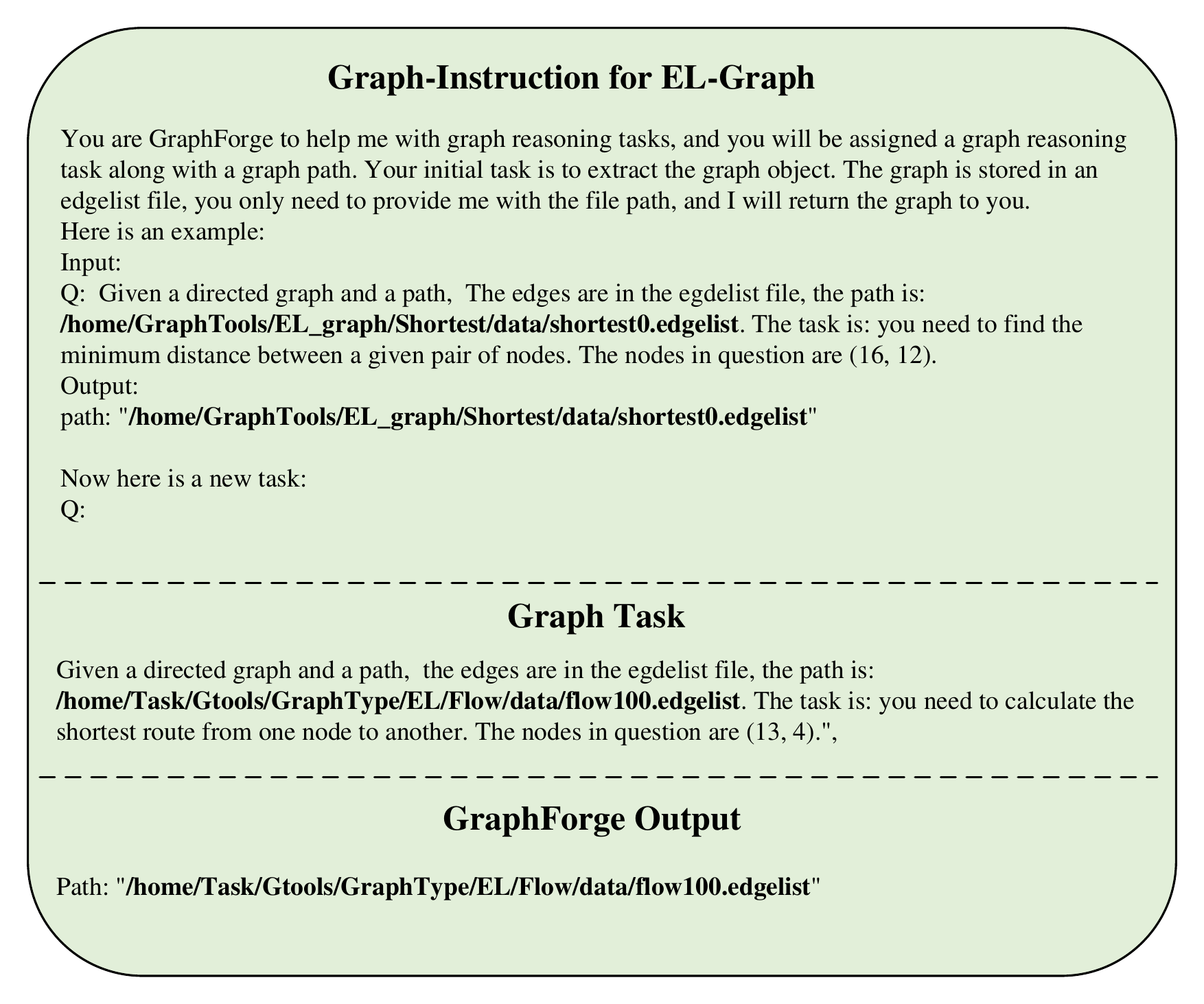}%%%%%%%%%%%%%%%%scale=缩小比例，或者用width=2in
\caption{Graph-Instruction for Exceeds Limit Graph (EL-Graph). Given the size constraints of EL-Graph, it becomes hard to extract complete graph structure information based on natural language. Therefore, we replace the graph with the file path. For EL-Graph, a one-shot prompt is used to direct LLMs to identify and extract the file path, enabling tools to retrieve graph structure information from the specified path.}  
\vspace{-1.5em}
\label{Fig: Graph-I-EL}
\end{figure*}

\begin{figure*}[t]
\centering 
\includegraphics[width=6.5in]{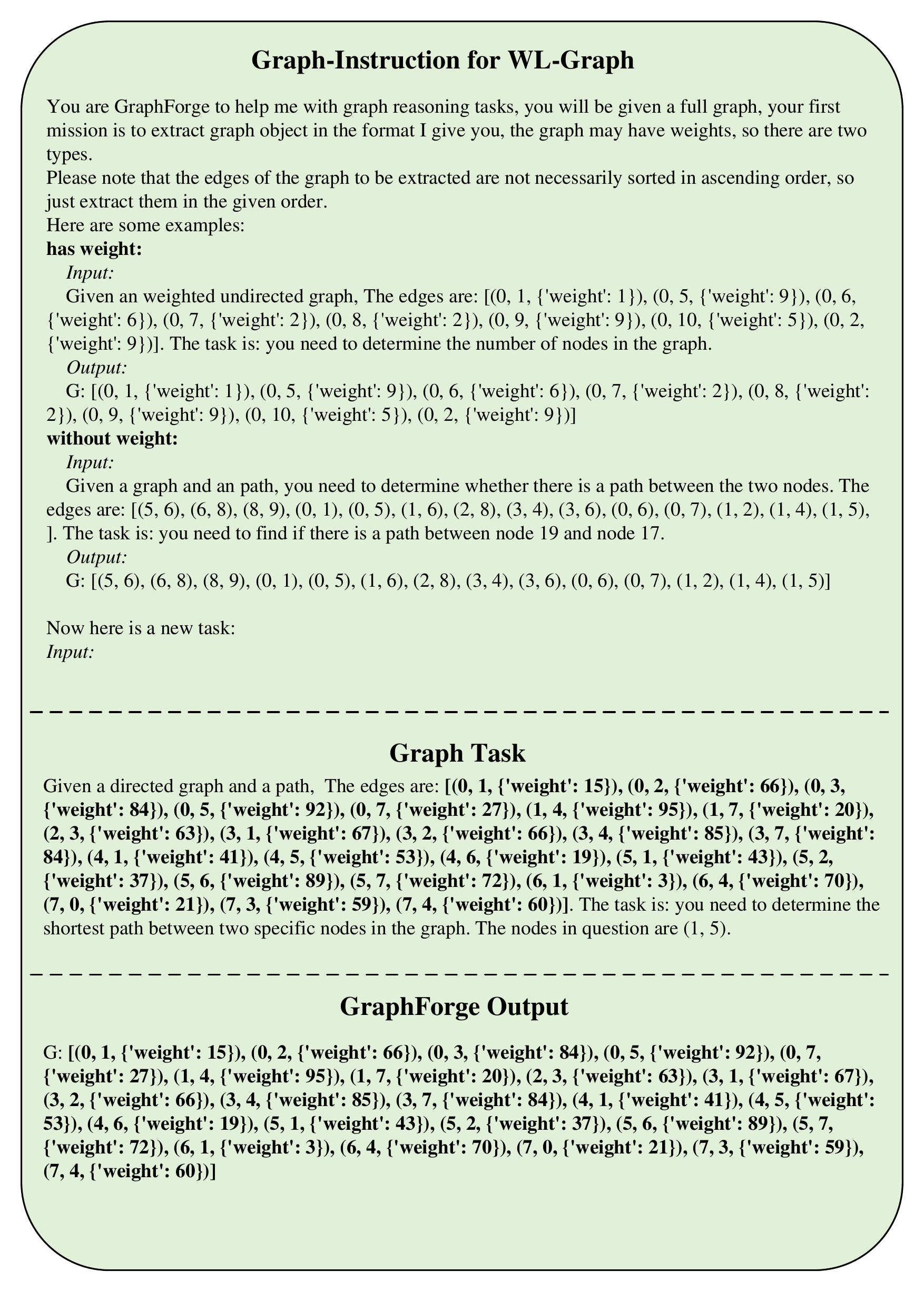}%%%%%%%%%%%%%%%%scale=缩小比例，或者用width=2in
\caption{Graph-Instruction for Within Limit Graph (WL-Graph), we employ a Two-shot Prompt of extracting graph information for both weighted and unweighted graphs in a list format of NetworkX.}  
\label{Fig: Graph-I-WL}
\end{figure*}

\begin{figure*}[t]
\centering 
\includegraphics[width=6.5in]{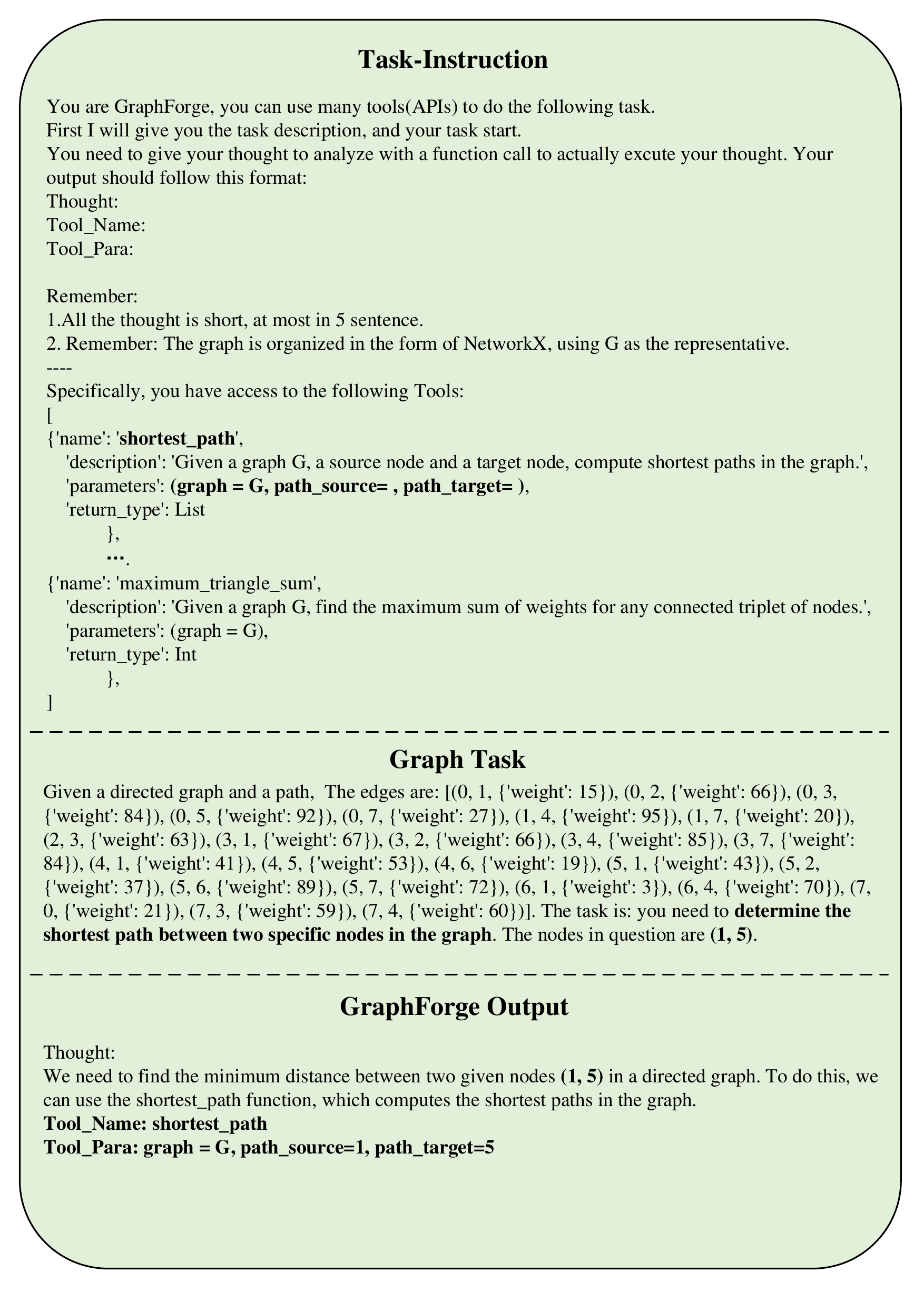}%%%%%%%%%%%%%%%%scale=缩小比例，或者用width=2in
\caption{The details of Task-Instruction. Based on a set of tools, Task-Instruction guides LLMs to choose the right graph tools for solving graph reasoning tasks, along with constraints on the output formats of tools.}  
\vspace{-1.5em}
\label{Fig: Task-I-WL}
\end{figure*}

\begin{figure*}[t]
\centering 
\includegraphics[width=6.5in]{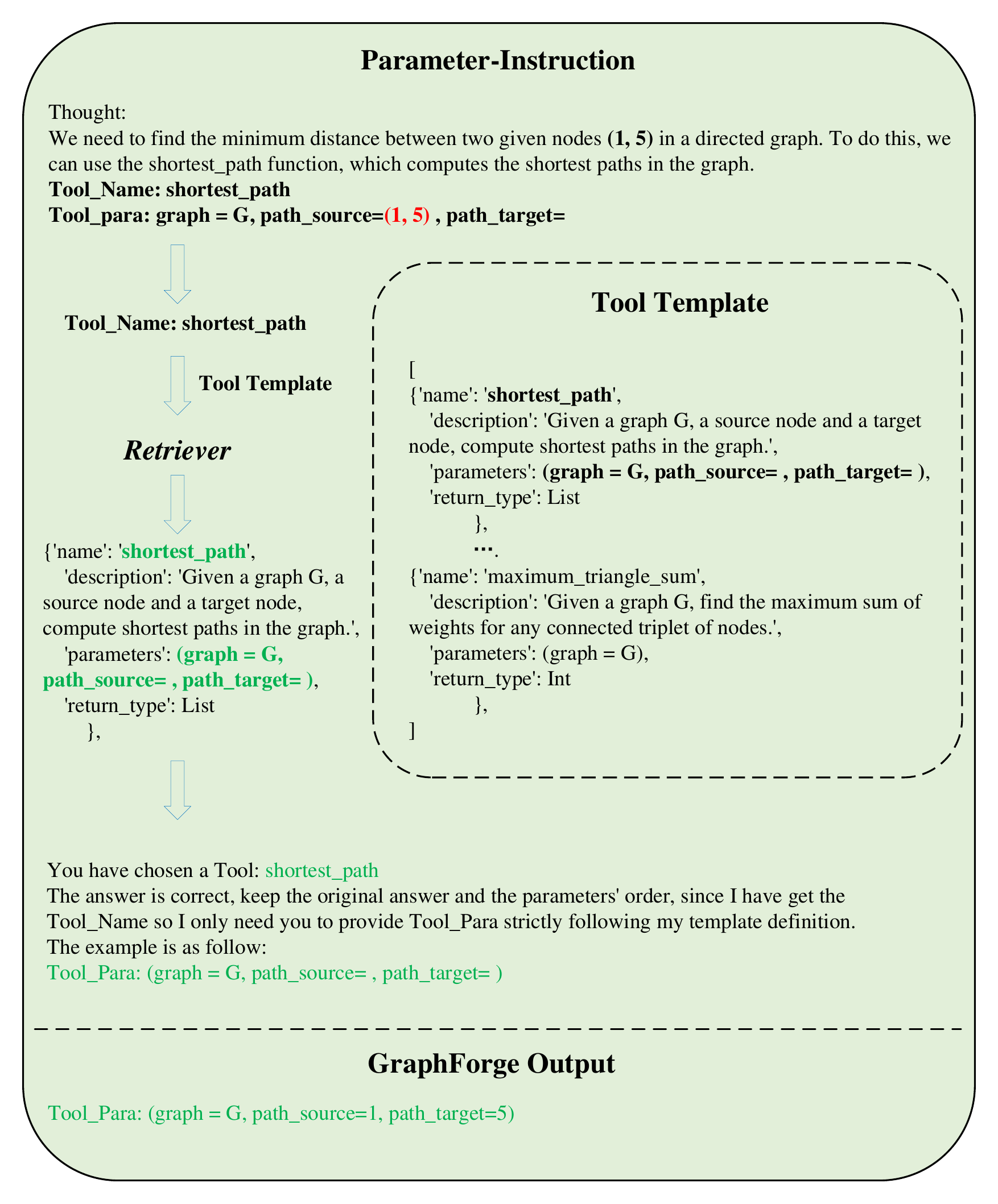}%%%%%%%%%%%%%%%%scale=缩小比例，或者用width=2in
\caption{The details of Parameter-Instruction. We first illustrate a possible error (\textit{e.g.}, an incorrect order causes Syntax Error) from GraphForge's output based on Task-Instruction. We define a Tool Template Retriever to retrieve the predefined format of parameters according to the Tool Name identified from Task-Instruction. Then we let GraphForge to extract the parameter based on Parameter-Instruction concatenated with the retrieval result.}  
\vspace{-1.5em}
\label{Fig: Para-I-WL}
\end{figure*}

Based on three components: Graph-Instruction, Task-Instruction and Parameter-Instruction. We exemplify the details of how GraphForge solves the corresponding  three subtasks: \textit{graph extraction} (in Figure~\ref{Fig: Graph-I-EL}, ~\ref{Fig: Graph-I-WL}), \textit{tool name identification} (in Figure~\ref{Fig: Task-I-WL}) and \textit{tool parameter extraction} (in Figure~\ref{Fig: Para-I-WL}).

\section{Error Analysis}
\label{appendxf Case study on error}
\begin{figure*}[t]
\centering 
\includegraphics[width=6.5in]{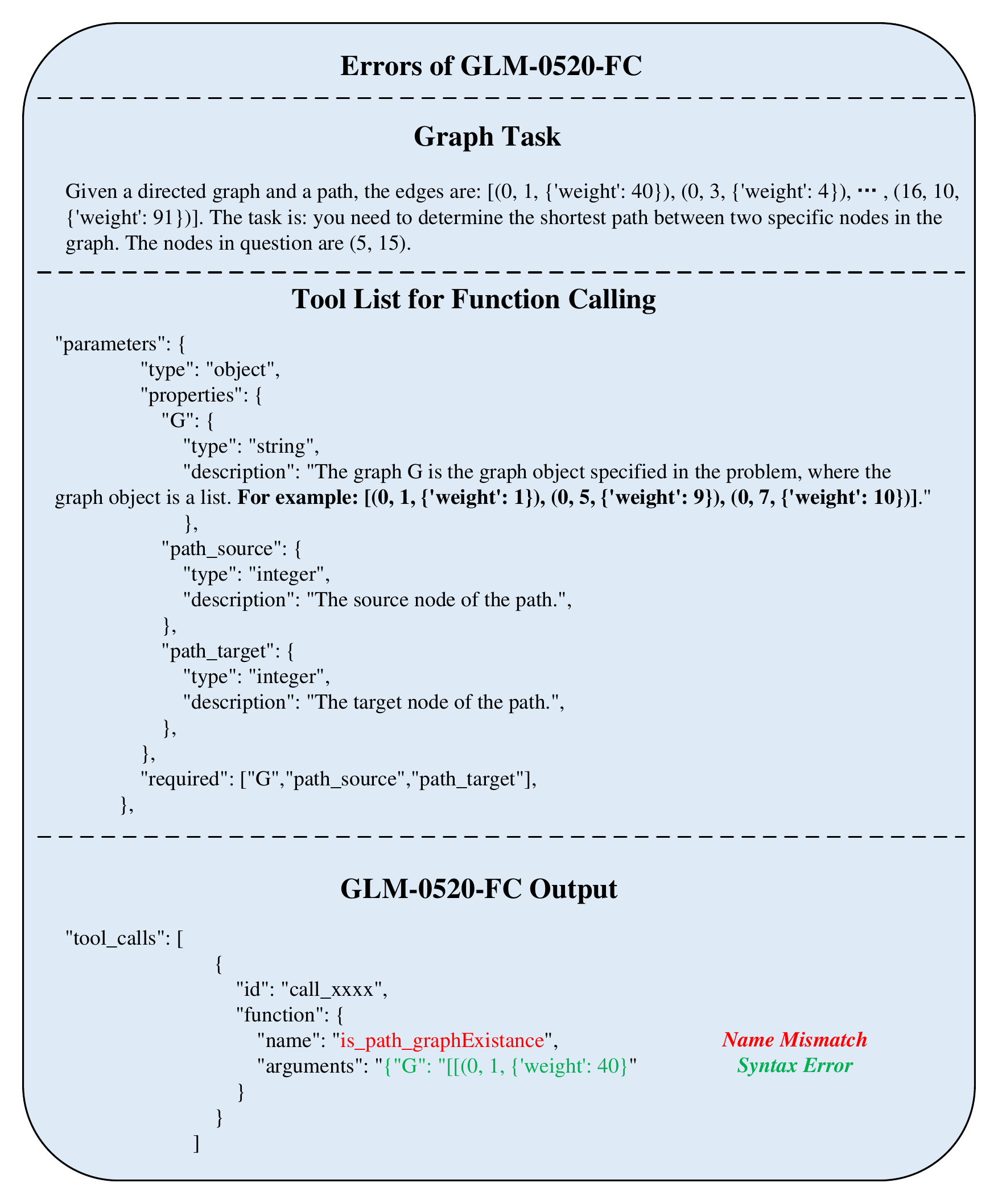}%%%%%%%%%%%%%%%%scale=缩小比例，或者用width=2in
\caption{Errors of GLM-0520-FC. For each LLM based on Function Calling, we provide the Tool List following their specified format. Additionally, to ensure the accuracy of graph information extraction, we set different prompts for graph extraction. This Figure shows both the Name Mismatch and Syntax Error happeneded in one case. Notably, we have found that GLM-0520-FC lacks the capability to process weighted graphs, as each result returned contains only a single edge. Consequently, we have made some modifications, as shown in Figure~\ref{Fig: Error GLM2}, specifically for GLM-0520-FC.}  
\vspace{-1.5em}
\label{Fig: Error GLM1}
\end{figure*}

\begin{figure*}[t]
\centering 
\includegraphics[width=6.5in]{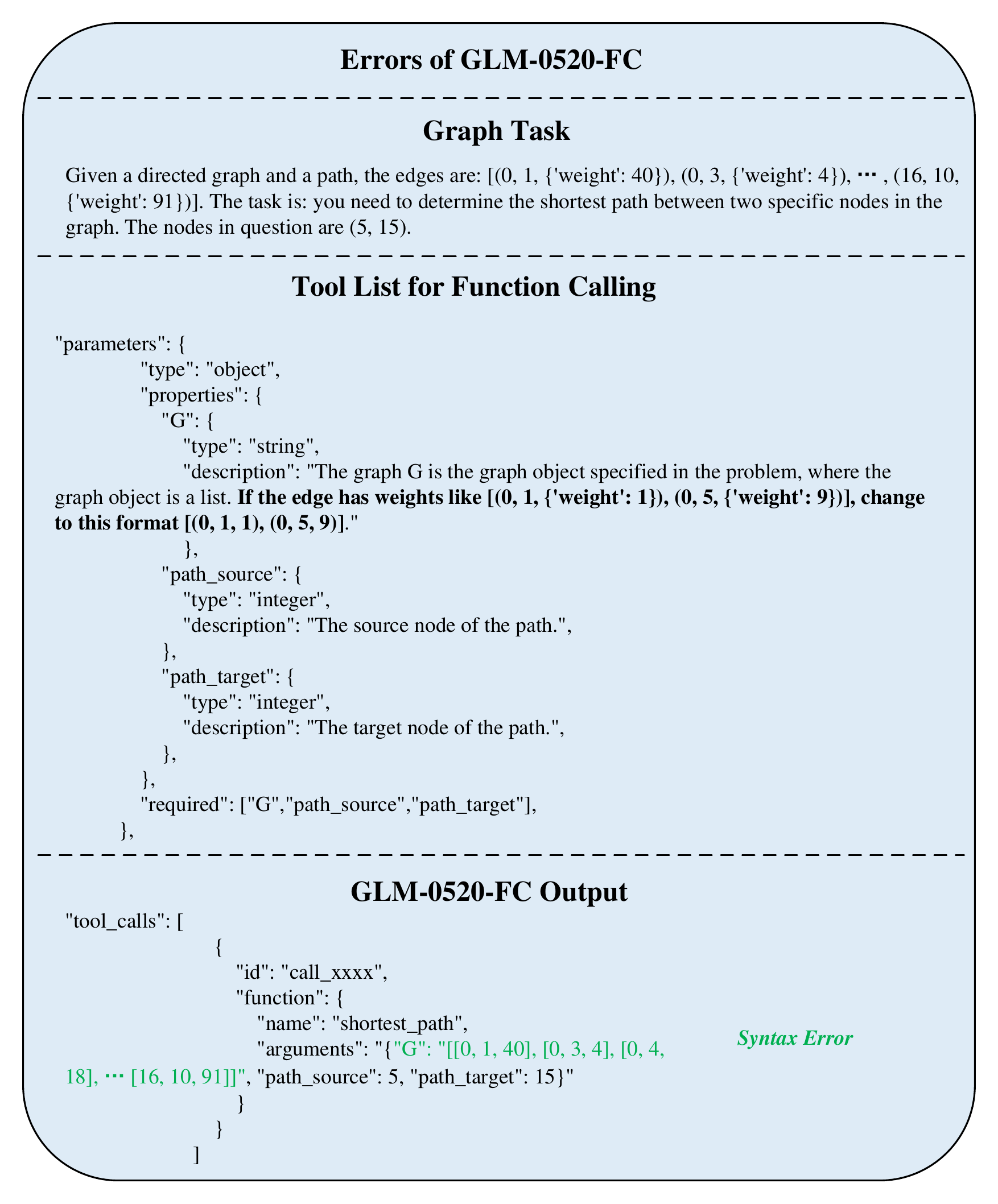}%%%%%%%%%%%%%%%%scale=缩小比例，或者用width=2in
\caption{Errors of GLM-0520-FC. Due to the case observed in Figure~\ref{Fig: Error GLM1}, we design a different prompt for GLM-0520-FC to convert weighted edges into triples. This alleviates the model's inability to recognize graph information, but Syntax Errors still occur.}  
\vspace{-1.5em}
\label{Fig: Error GLM2}
\end{figure*}

\begin{figure*}[t]
\centering 
\includegraphics[width=6.5in]{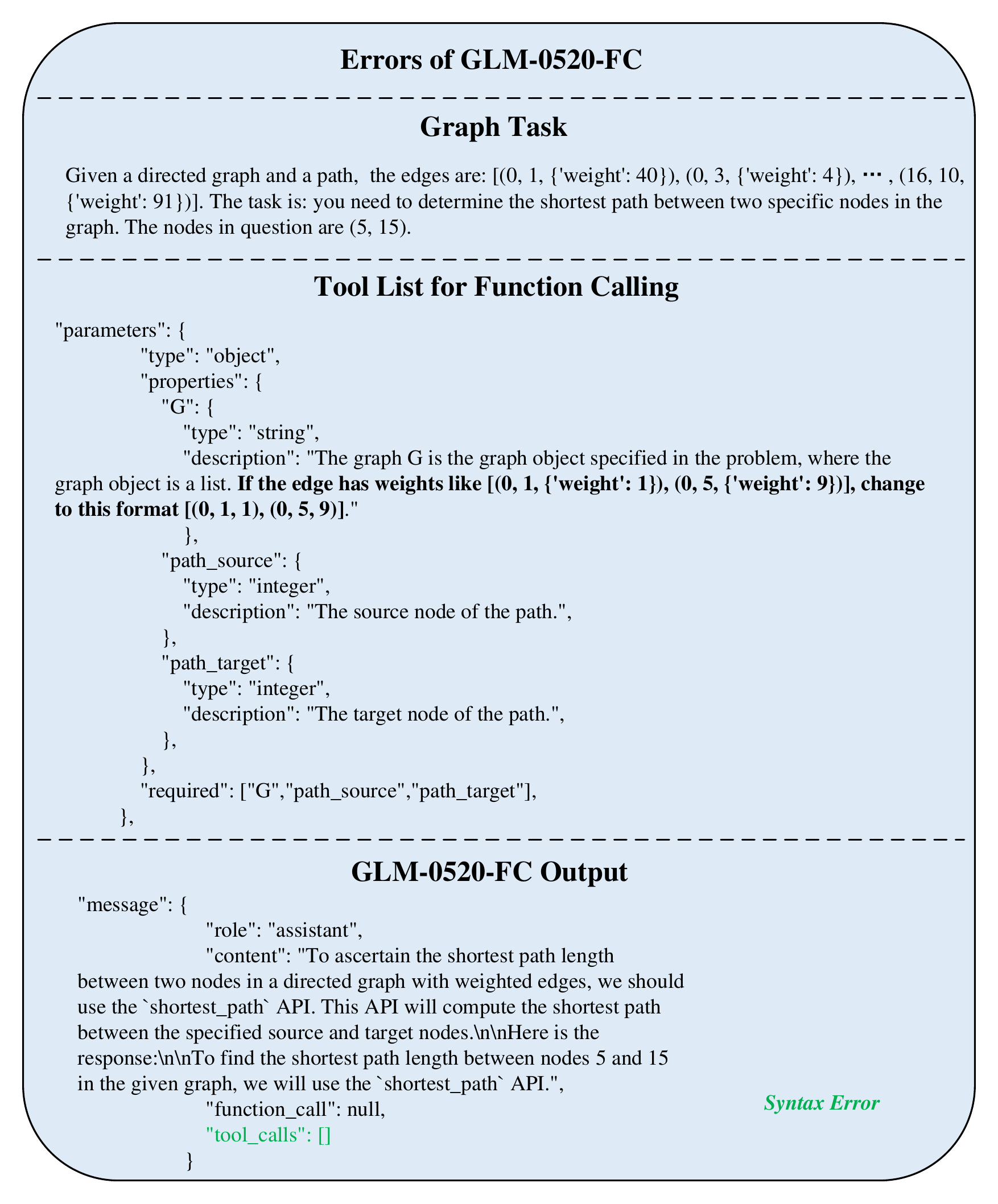}%%%%%%%%%%%%%%%%scale=缩小比例，或者用width=2in
\caption{Errors of GLM-0520-FC. Apart from above Syntax Errors, we have found some randomly occurring instances of missing parameters. By observing the content returned by the API, we analyze that the model ignores graph information during the extraction process, resulting in no return results.}  
\vspace{-1.5em}
\label{Fig: Error GLM3}
\end{figure*}

\begin{figure*}[t]
\centering 
\includegraphics[width=6.5in]{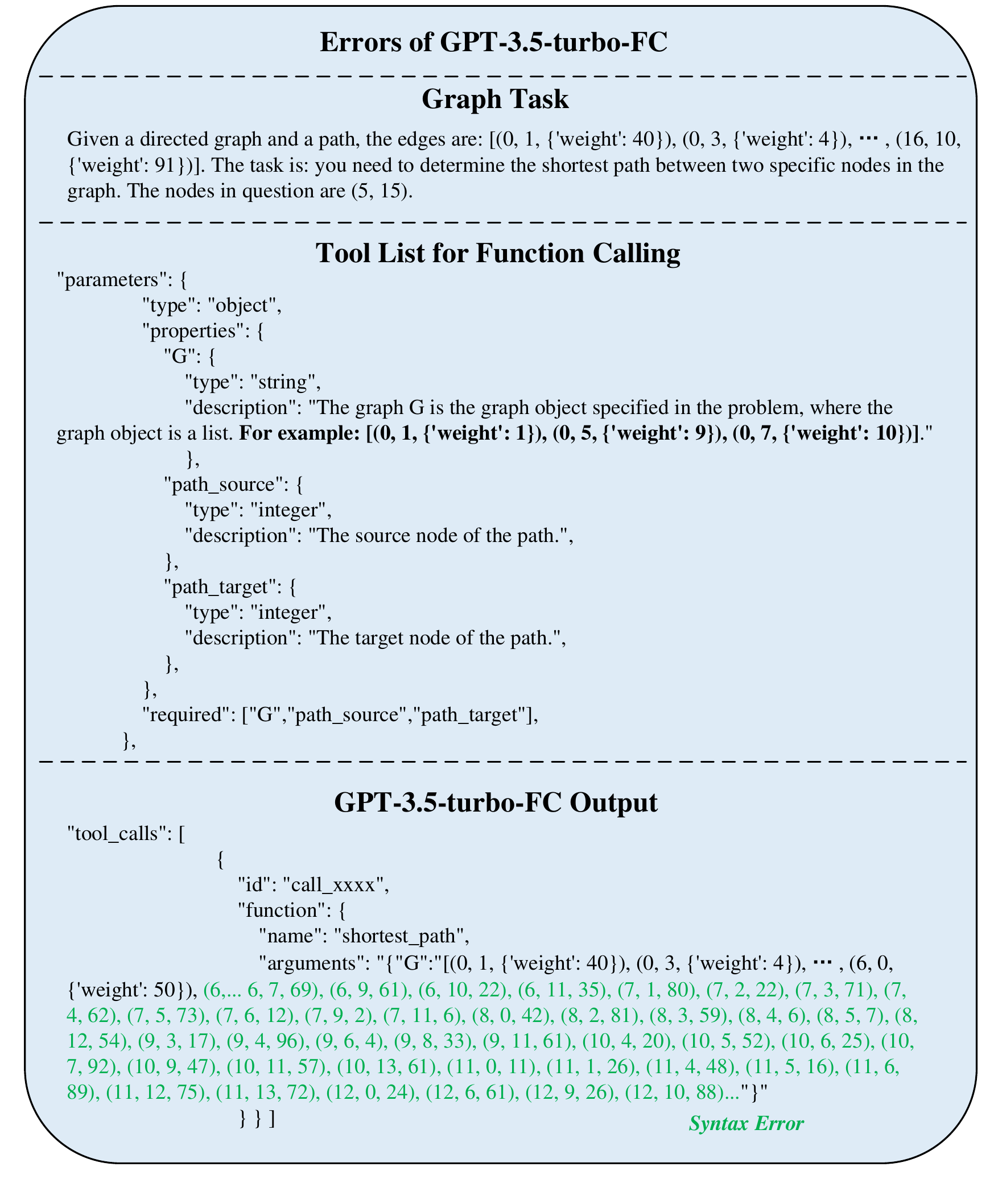}%%%%%%%%%%%%%%%%scale=缩小比例，或者用width=2in
\caption{Errors of GPT-3.5-turbo-FC. For each LLM based on Function Calling, we provide the Tool List following their specified format. Additionally, to ensure the accuracy of graph information extraction, we set different prompts for graph extraction. However, we still find some randomly occurring instances of missing parameters, which cause Syntax Errors.}  
\vspace{-1.5em}
\label{Fig: Error GPT}
\end{figure*}

\begin{figure*}[t]
\centering 
\includegraphics[width=6.5in]{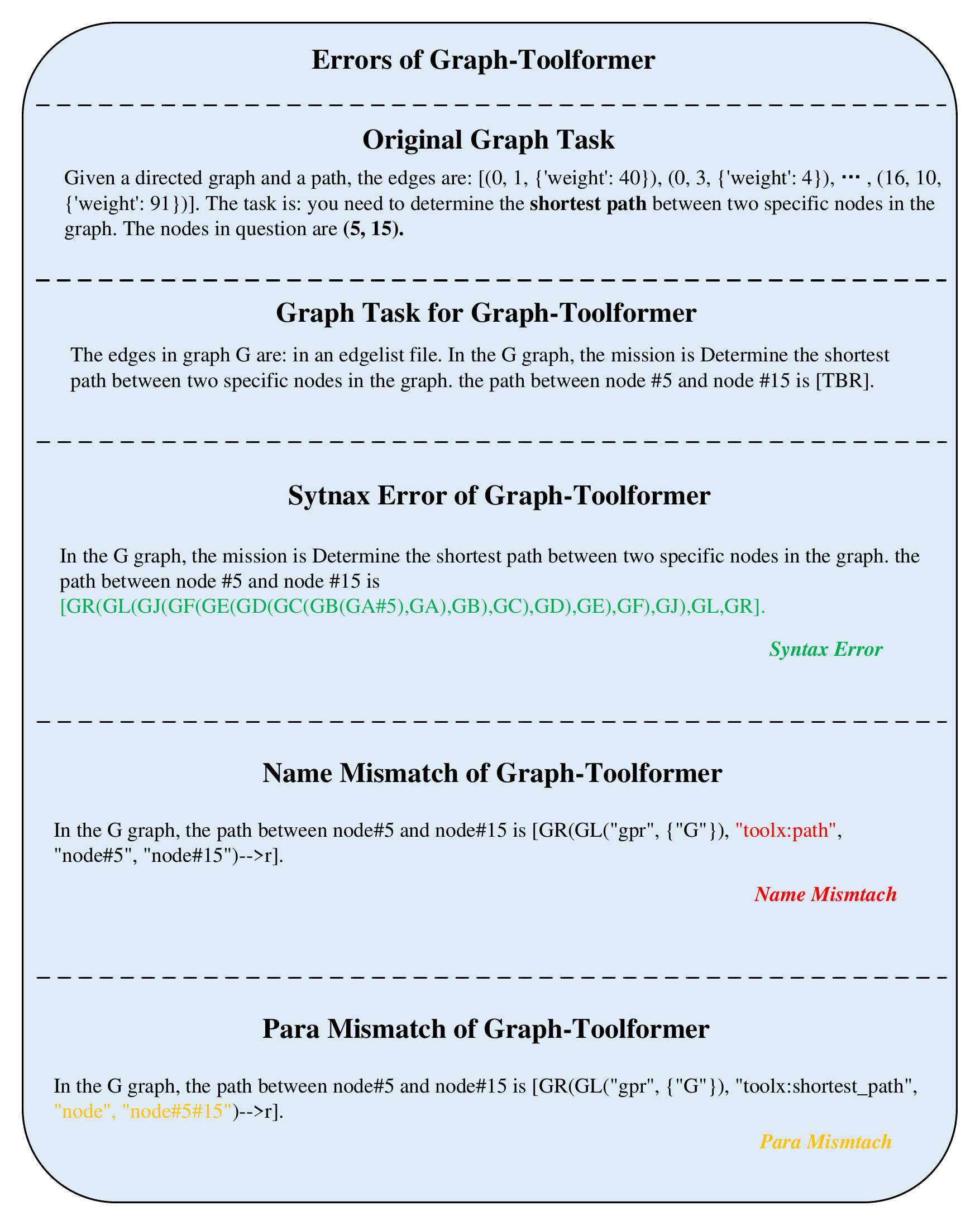}%%%%%%%%%%%%%%%%scale=缩小比例，或者用width=2in
\caption{Errors of Graph-Toolformer. We transform the original task into a special instruction format for Graph-Toolformer. When checking the output, we have found various errors associated with Graph-Toolformer.}  
\vspace{-1.5em}
\label{Fig: Error GT}
\end{figure*}

\begin{figure*}[t]
\centering 
\includegraphics[width=6.5in]{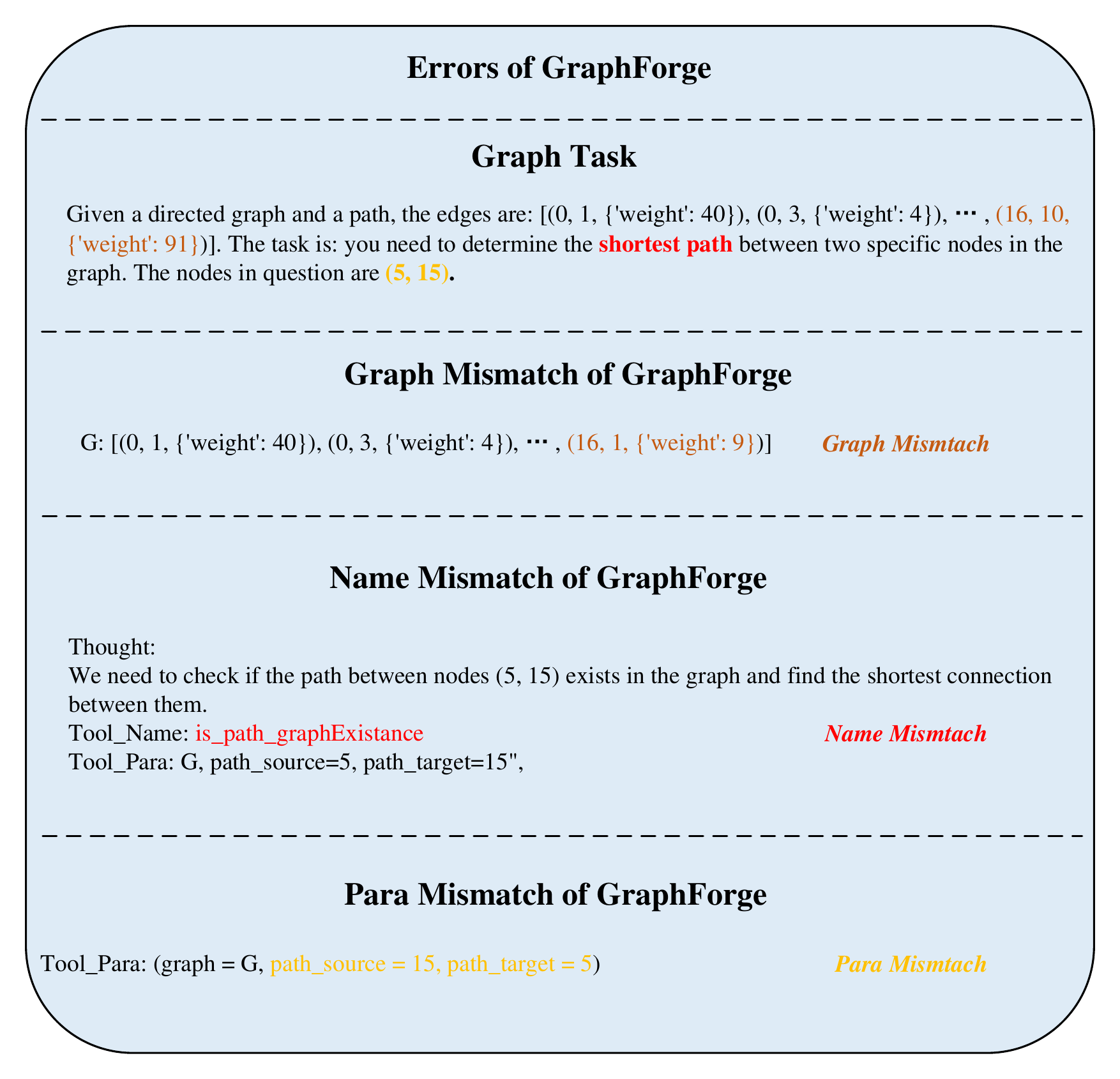}%%%%%%%%%%%%%%%%scale=缩小比例，或者用width=2in
\caption{Errors of GraphForge. We have found that Graph Mismatch may occur when the input content is very long. Additionally, there is a small probability that GraphForge will have a tool misidentification, leading to a Name Mismatch. Before we implement the Parameter-Instruction, the model frequently exhibits incorrect order of parameters. This issue has almost disappeared after applying the Parameter-Instruction.}  
\vspace{-1.5em}
\label{Fig: Error GF}
\end{figure*}

In this section, we present error cases about the Tool-Instruction methods and our method. Figure~\ref{Fig: Error GLM1}, ~\ref{Fig: Error GLM2}, ~\ref{Fig: Error GLM3} present the errors of GLM-0520-FC. It should be noted that due to GLM-0520-FC's inability to extract graph edges with weights, we use a one-shot prompt to let GLM-0520-FC output edge list in a triplet form. Figure~\ref{Fig: Error GPT} shows the omission of the graph edges and tool parameters from GPT-3.5-turbo which causes Syntax Errors. Figure~\ref{Fig: Error GT} shows the Mismatch and Syntax Error of Graph-Toolformer. Figure~\ref{Fig: Error GF} shows some errors of GraphForge.

\end{document}